\documentclass[10pt,twocolumn,letterpaper]{article}

\usepackage{cvpr}              

\definecolor{cvprblue}{rgb}{0.21,0.49,0.74}
\usepackage[pagebackref,breaklinks,colorlinks,allcolors=cvprblue]{hyperref}
\usepackage[accsupp]{axessibility} 
\usepackage[utf8]{inputenc} 
\usepackage[T1]{fontenc}    
\usepackage{url}            
\usepackage{booktabs}       
\usepackage{amsfonts}       
\usepackage{nicefrac}       
\usepackage{microtype}      
\usepackage{xcolor}         
\usepackage{amssymb}
\usepackage{pifont}
\usepackage{graphicx} 
\usepackage{multirow}
\usepackage{amsmath} 
\usepackage{adjustbox}
\usepackage{xspace}
\newcommand{\cmark}{\ding{51}}%
\newcommand{\xmark}{\ding{55}}%
\setcitestyle{numbers,square}

\newcommand{\method}{VGGHeads\xspace}

\def\paperID{*****} 
\def\confName{CVPR}
\def\confYear{2026}

\title{\method: 3D Multi Head Alignment with a Large-Scale Synthetic Dataset}

\author{
Orest Kupyn\textsuperscript{1,2} \ 
Eugene Khvedchenia\textsuperscript{3} \ 
Christian Rupprecht\textsuperscript{1} \\  
\\
\textsuperscript{1} University of Oxford \
\textsuperscript{2} Piñata Farms, Los Angeles, USA \
\textsuperscript{3} Ukrainian Catholic University \ 
}

\begin{document}
\twocolumn[{%
\renewcommand\twocolumn[1][]{#1}%
\maketitle

\begin{center}
    \centering
    \captionsetup{type=figure}
    
    \includegraphics[width=\textwidth]{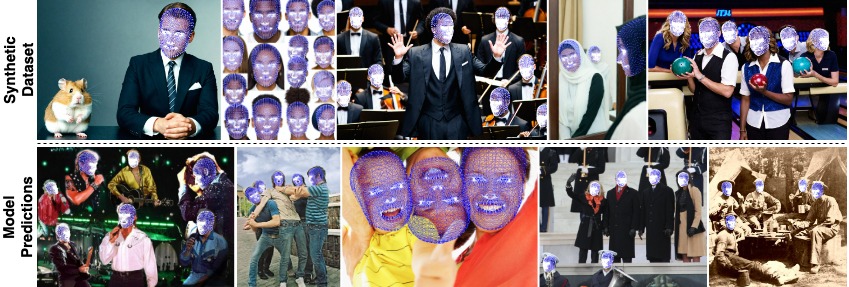}

    \caption{\textbf{\method} Top: Our fully synthetic large-scale dataset for 3D Multi Head Alignment with a wide variety
of scenes, numbers of people, and rich annotations for
every human head. Bottom: The model trained on this fully synthetic data generalizes to in-the-wild scenes.}
\end{center}

}
]

\begin{abstract}
Human head detection, keypoint estimation, and 3D head model fitting are essential tasks with many downstream applications. However, recent progress has been limited, as real-world detection datasets face significant privacy and ethical barriers, while existing crop-based datasets are unable to capture the multi-person scenarios required for real-world applications. Here, we introduce \method---a large-scale synthetic dataset of over 1 million multi-person images generated with controllable diffusion models, providing comprehensive 3D annotations for every head while significantly reducing privacy concerns. Using this dataset, we train a novel unified architecture that performs simultaneous head detection and coarse 3D mesh estimation from unconstrained images in a single forward pass. Extensive evaluations demonstrate that models trained exclusively on synthetic data achieve state-of-the-art performance on real-world benchmarks, proving that synthetic datasets can match the effectiveness of traditional data collection approaches
Furthermore, the versatility of our dataset provides a robust foundation for downstream detailed reconstruction methods while enabling diverse head modeling tasks. 
\end{abstract}
\vspace{-0.5cm}
\section{Introduction}

The demand for high-quality datasets has surged across computer vision, particularly for tasks involving human head representations. Accurate head modeling is crucial for applications from facial recognition \cite{arcface} and animation \cite{gauss_avatar} to augmented reality and medical imaging. Traditional datasets often fall short, focusing narrowly on specific aspects, such as facial landmarks \cite{face_align}, or offering limited resolution and annotation types.

Head processing traditionally follows a multi-stage pipeline: face detection, landmark detection, and 3D alignment on cropped regions. This sequential approach has fundamental limitations. Face detection operates within limited pose ranges and fails to capture the necessary variability for accurate 3D modeling. Importantly, a significant gap exists between coarse detection and fine-grained reconstruction, requiring practitioners to bridge representation levels through separate, specialized models.

Severe data limitations compound methodological challenges. Established datasets, such as VGG-Face \cite{vggface}, FFHQ \cite{ffhq}, and CelebA \cite{celeba}, assume tightly cropped, pre-aligned single faces under controlled conditions. While enabling advances in face recognition and generation, they cannot address multi-person, unconstrained scenarios, prevalent in real-world applications. Privacy and ethical challenges further hold back progress. Unlike some other computer vision domains, facial data requires explicit consent, creating fundamental barriers. Major benchmarks have been withdrawn: MS-Celeb-1M \cite{ms_celeb}, VGG-Face \cite{vggface}, and DukeMTMC \cite{dukemtmc} due to consent violations, while ImageNet \cite{imagenet} required face blurring in 2021 \cite{blurred_imagenet}.
These limitations create a dual bottleneck: existing methods cannot effectively bridge the detection-reconstruction gap, while privacy constraints prevent the collection of diverse, well-annotated datasets needed for unified models. Synthetic data generation offers a unique solution, enabling diverse multi-person scenarios with perfect annotations while eliminating privacy concerns.

We introduce \method, bridging coarse head detection and parametric 3D modeling through end-to-end learning. Our unified model processes unconstrained images and outputs complete 3D head representations for all visible people in a single forward pass, eliminating the need for separate detection, cropping, and reconstruction stages while jointly optimizing from bounding boxes to parametric 3D mesh parameters.
To enable this approach, we construct a large-scale synthetic dataset using latent diffusion models conditioned on 2D body skeletons from real image collections. Our dataset comprises over 1 million multi-person scenes with comprehensive 3D annotations for every head. Conditioning on 2D body skeletons effectively models diverse in-the-wild scenes while eliminating privacy concerns.

\noindent{Our contributions are as follows:}

\textbf{Unified Multi-Scale Head Representation:} We present the first end-to-end real-time model that bridges the gap between coarse head detection and fine-grained 3D reconstruction, jointly optimizing bounding boxes, 3D vertices, pose, and landmarks in a single forward pass through knowledge distillation from sequential pipelines, demonstrating superior performance over the original teacher while eliminating separate cropping and alignment stages.

\textbf{Large-Scale Multi-Person Synthetic Dataset:} We introduce a comprehensive dataset of over 1 million images with 2.2 million annotated heads, with each head annotated with a detailed 3D head mesh. Unlike existing cropped and pre-aligned datasets, our data contains multiple faces per frame. Compared to detection benchmarks, it is over 20 times larger than WIDERFace \cite{wider} (32,303 images), while providing substantially more information per frame. The detailed representation enables various downstream tasks from face detection to 3D alignment, while generation with diffusion models \cite{sdxl} eliminates privacy concerns.

\textbf{Synthetic-to-Real Generalization:} We demonstrate that models trained exclusively on synthetic data achieve state-of-the-art performance on real-world benchmarks, establishing that synthetic datasets can exceed the traditional real-data approaches. This proves the viability of privacy-preserving synthetic data pipelines for advancing computer vision research.
\section{Related Work}

\begin{figure*}[t!]
\includegraphics[width=0.99\textwidth]{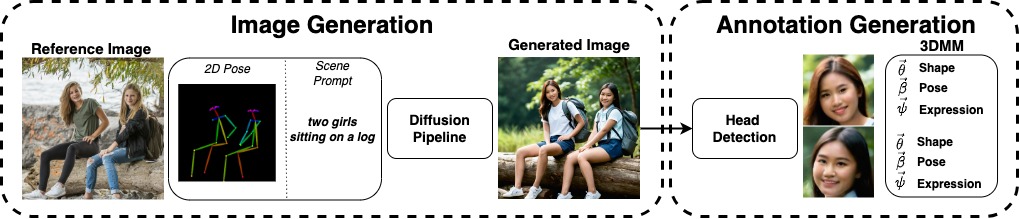}
\centering
\caption{\textbf{Data Generation.} The predicted 2D body pose \cite{yolonas} and scene description \cite{blip} condition the image generation process. Binary detection model predicts head bounding boxes and 3DMM regressor \cite{dad3d} generates final annotation for each head crop.}
\label{fig:architecture}
\end{figure*}

\textbf{Head Detection:} The simplest head representation is a face bounding box. Early detectors used hand-crafted features \cite{hog_face, viola_face}, but deep learning significantly reshaped the field \cite{fasterrcnn}. Current methodologies are categorized into single-stage \cite{yolo_review, retina_net} and two-stage methods \cite{fasterrcnn}. Single-stage approaches like S$^3$FD \cite{s3fd} and PyramidBox \cite{pyramidbox} efficiently detect smaller facial structures. Two-stage methods based on Faster R-CNN \cite{fasterrcnn} and R-FCN \cite{rfcn} integrate multi-scale techniques \cite{fdnet, facerfcn}. However, these approaches are limited to facial areas and struggle with side poses and out-of-distribution scenarios.
HollywoodHeads \cite{hollywood_heads} extends to full head detection with movie scenes, while SCUT-Head \cite{scut-head} introduces 4,405 classroom images with cascade detection. Both datasets are limited to specific scenes, not reflecting real-world distributions.
RetinaFace \cite{retinaface} improved accuracy through multi-task learning of bounding boxes and facial landmarks, taking a first step toward general head representation. img2pose \cite{img2pose} directly regresses 6DoF face pose, recovering bounding boxes and 3D head pose. However, these methods only predict pose or limited facial landmarks, insufficient for comprehensive head modeling.

\textbf{3D Morphable Model:} Early 3DMMs provided general-purpose head representations \cite{3dmm}—statistical models of 3D facial shapes and textures from scanned faces enabling realistic 3D model creation with few parameters. The Basel Face Model \cite{bazelface} uses matrix decomposition with parameters from 200 scans. Recent models like FLAME \cite{FLAME} cover full 3D head meshes trained on larger datasets.
Several methods address regressing 3DMM parameters from head crops. RingNet \cite{RingNet} estimates 3D face shape without direct 3D supervision. 3DDFA \cite{3ddfa_cleardusk} uses Cascaded CNNs for dense 3DMM prediction, while 3DDFA-v2 \cite{3ddfav2} improves through meta-joint optimization. DAD-3DHeads \cite{dad3d} introduces the first in-the-wild 3D head dataset with direct supervision. DECA \cite{deca}, EMOCA \cite{emoca}, and MICA \cite{mica} address finer geometry modeling but require robust detection and alignment for high-quality crop extraction.
\method bridges this gap, providing robust multi-person head detection and coarse 3D estimates that serve as reliable initialization for detailed reconstruction approaches. RetinaFace \cite{retinaface} first recovers 3D landmarks end-to-end but is limited to facial areas and uses only 5 ground-truth 2D landmarks for pseudo 3D information, lacking consistency on challenging samples. PIXIE \cite{pixie} and PyMAF-X \cite{pymaf} attempt joint head and body reconstruction but model limited parameters, lack robustness on out-of-distribution poses, and rely on separate coarse predictions.

Lack of large-scale datasets for direct head mesh regression and task complexity due to high-dimensional representations limits progress. \method addresses these issues with a large-scale synthetic dataset containing dense 3D annotations and, to our knowledge, the first model to directly recover multi-person head meshes from images.

\textbf{Synthetic Data Generation:} Early synthetic data generation used 3D rendering engines for 2D vision problems. These approaches were constrained by 3D model domains and required modifications for each dataset and subtask. Virtual KITTI \cite{virtual_kitti} is limited to street driving scenes. Recent methods using generative adversarial networks (GANs) \cite{synth_saliency} offer greater flexibility and generalization. However, these primarily sample from original dataset distributions, limiting their ability to incorporate new information. Diffusion-based methods like Diffumask \cite{diffumask} and DatasetDM \cite{datasetdm} generate synthetic images and annotations for semantic segmentation and depth prediction. Instance Augmentation \cite{instance_aug} generates separate objects in images, providing frameworks to augment and anonymize datasets. However, these methods require fine-tuning for specific datasets and have limited application to tasks with low ground truth data.

\textbf{Synthetic Face Data:} Synthetic face generation typically uses rendering engines. Fake It Till You Make It \cite{fakeit} releases the first large-scale synthetic dataset combining procedurally generated 3D face models with hand-crafted assets to render training images with 3D head meshes and 2D landmarks. However, the dataset contains only tight face crops and requires manual mesh texture generation, limiting scalability.
Alternative approaches manipulate 2D images instead of 3D graphics pipelines. Some methods \cite{face_align} fit 3DMMs to face images and warp them for head pose augmentation, while others \cite{hand2face} composite hand images onto faces to improve detection. These approaches make only minor adjustments to existing images, limiting utility.
In contrast, \method directly samples from large diffusion model distribution, easily scaling to arbitrary numbers of training samples and scene types while capturing diverse multi-person scenarios impossible with crop-based approaches.
\section{Dataset}

\begin{table*}
\centering
\setlength{\tabcolsep}{3pt}
\footnotesize
\caption{\textbf{Dataset Comparison.} Overview of prominent face/head datasets showing scale, multi-person capability, and annotation types. \method combines large scale with comprehensive multi-person 3D annotations, enabling unified multi-task head modeling.}\label{t:dataset_comp}
\resizebox{\textwidth}{!}{%
\begin{tabular}{@{}lccccccc@{}}
\toprule
\textbf{Dataset} & \textbf{Images} & \textbf{Heads} & \textbf{Multi-Person} & \textbf{Head Box} & \textbf{Face Box} & \textbf{3D Landmarks} & \textbf{Status/Issues} \\
\midrule
DAD‑3DHeads \cite{dad3d} & 44,898 & 44,898 & \xmark & \cmark & \cmark & \cmark (5,023) & Cropped \& Aligned \\
CelebA \cite{celeba} & 202,599 & 202,599 & \xmark & \xmark & \cmark & \xmark & Cropped \& Aligned \\
VGG-Face \cite{vggface} & 2,600,000 & 2,600,000 & \xmark & \xmark & \cmark & \xmark & \textcolor{red}{Withdrawn} \\
MS-Celeb-1M \cite{ms_celeb} & 10,000,000 & 10,000,000 & \xmark & \xmark & \cmark & \xmark & \textcolor{red}{Withdrawn} \\ \midrule
FDDB \cite{fddb} & 2,845 & 5,171 & \cmark & \cmark & \xmark & \xmark & Limited scenes \\
WIDER FACE \cite{wider} & 32,203 & 393,703 & \cmark & \xmark & \cmark & \xmark & Face Labels Only \\
HollywoodHeads \cite{hollywood_heads} & 224,740 & 369,846 & \cmark & \cmark & \xmark & \xmark & Limited scenes \\
\midrule
\textbf{\method (Ours)} & \textbf{1,022,944} & \textbf{2,219,146} & \cmark & \cmark & \cmark & \cmark \textbf{(5,023)} & \textcolor{ForestGreen}{Synthetic/Private} \\
\bottomrule
\end{tabular}%
}
\end{table*}

Our goal is to create a large-scale dataset of image-label pairs where each image contains multiple people and every visible head is annotated with precise bounding boxes and complete 3D morphable model parameters. Unlike existing crop-based datasets assuming pre-aligned single faces, our dataset captures unconstrained multi-person scenarios essential for real-world applications. We generate fully synthetic images using pretrained latent diffusion models \cite{sdxl}, then employ multi-stage annotation and filtering to ensure high-quality 3D annotations.

\subsection{Generation Pipeline}
The dataset generation process consists of the following stages: (1) images are generated with a latent diffusion model conditioned on a large real-world dataset, (2) a small subset of data is manually labeled with head bounding boxes to train a binary head detector on synthetic data, (3) for each detected head in generated images we predict the 3D head model parameters and (4) multi-level quality filtering ensures dataset reliability. The full pipeline is illustrated in \cref{fig:architecture}.

\textbf{Image Generation:} Latent diffusion models have achieved remarkable progress in image generation \cite{sdxl, cascadediff}. However, generating diverse multi-person scenes with coherent spatial compositions remains challenging \cite{diffumask}. To create a dataset that generalizes to in the wild settings, we require scenes with varied backgrounds, different numbers of people, and complex inter-person interactions. 
We address this through controllable generation using 2D human pose skeletons as an intermediate representation. Body poses encode positions, scales, and spatial relationships of multiple people, providing the structural control necessary for reliable multi-person scene generation. We condition image generation on poses extracted from LAION \cite{laion} using pose estimation \cite{yolonas}, ensuring our synthetic scenes reflect real-world spatial distributions. To eliminate privacy concerns, we use BLIP-VQA \cite{blip} to generate subject-neutral scene descriptions rather than using original captions that may contain identifying information. This approach also enables bias reduction by modifying prompts to ensure diverse demographic sampling.
The T2I-Adapter \cite{t2i} injects skeleton maps into SDXL \cite{sdxl} decoder blocks, generating high-resolution synthetic images up to 1280×1280 pixels with precise control over human placement and pose diversity.

\textbf{Annotation Process:} Existing face detectors \cite{retinaface} are optimized for frontal regions and fail on extreme poses and multi-person scenarios. We manually annotate 10,000 uniformly selected images with full head bounding boxes—the smallest axis-parallel rectangles encompassing all visible head pixels, including hair and occlusions.
This serves dual purposes: training a robust binary head detector and establishing annotation standards. We remove harmful or privacy-sensitive content during manual review. RT-DETR \cite{rtdetr} trained on this subset achieves \textbf{0.925 mAP} on 2000 validation images, demonstrating feasibility of training robust detectors on synthetic data.
Using our detector, we locate and crop all heads for 3D annotation. We employ FLAME \cite{FLAME}, which parameterizes head geometry using 413 disentangled parameters covering shape, expression, and pose. The state-of-the-art FLAME regressor \cite{dad3d} generates comprehensive annotations including mesh vertices, pose rotations, and expression coefficients.

\textbf{Multi-Stage Quality Filtering:} Despite advances in diffusion models, generation failures can occur, particularly in complex multi-person scenes. We implement a comprehensive four-stage filtering pipeline specifically designed to identify and remove problematic samples while maintaining high recall of valid data.

\textit{Basic Validity:} Remove images with zero detected heads, indicating complete generation failure.

\textit{Consistency Verification:} Apply head detection to both original and horizontally flipped images. Remove samples where head counts differ, indicating unstable detections or generation artifacts.

\textit{Cross-Method Validation:} Verify that our head detector's outputs are consistent with face detection using RetinaFace \cite{retinaface}. Remove images where detected faces have no spatial overlap with detected heads, catching cases where head detection fails while face detection succeeds.

\textit{Scale Consistency:} Split images vertically and verify that the sum of heads detected in both halves equals the total in the original image. This removes images with many tiny, often deformed heads due to resolution limitations.

Our pipeline achieves \textbf{97.3\% recall}, treating minor artifacts as beneficial data augmentation given our coarse task nature. This ensures dataset scale while guaranteeing task-relevant quality.

\begin{figure*}[t!]
\includegraphics[width=0.99\textwidth]{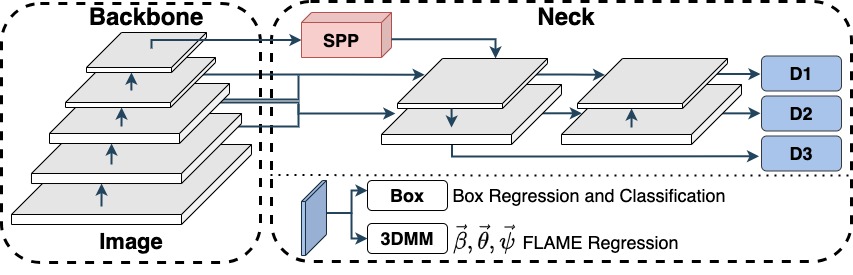}
\centering
\caption{\textbf{Model Architecture.} \method extends YOLO-NAS \cite{yolonas} architecture to predict the 3D Morphable Model parameters along with the head bounding boxes from the multi-scale feature maps (D1-D3).}
\label{fig:model_arch}
\end{figure*}

\subsection{Safety \& Privacy}

Traditional face datasets face significant privacy challenges, with major benchmarks withdrawn due to consent violations. Our synthetic approach eliminates these concerns while enabling unlimited scalability.

\textbf{Content Safety:} We implement multi-layered harmful content prevention: filter source data, employ CLIP-based NSFW filtering \cite{nsfw} with high sensitivity, incorporate negative embeddings for problematic keywords, and apply open-source NSFW classification \cite{nsfw_classifier} achieving \textbf{99.49\% recall} and \textbf{83.37\% precision} on 2,101 manually verified images.

\textbf{Content Privacy:} Controlled experiments fine-tuning Stable Diffusion on RaFD \cite{rafd} with 61 individuals show that multi-person training does not enable identity memorization and re-identification models \cite{arcface} fail to match generated faces to training individuals. However, single-person fine-tuning enables memorization, suggesting identity information is stored in text encoders linked to names.
We mitigate residual risks using GliNER \cite{gliner} to detect and remove personal names, BLIP \cite{blip} for privacy-neutral descriptions, and filtering against CelebA \cite{celeba} celebrity databases. This multi-component approach eliminates privacy concerns causing major dataset withdrawals.

\subsection{Dataset Statistics} 
From 1.7 million LAION-FACE \cite{laion_face} images, we filter 20.6\% for privacy/NSFW content, retaining only pose skeletons and neutral descriptions while discarding visual data. After generating 1.3 million synthetic images and applying quality filtering (removing 19.9\%), our final dataset contains \textbf{1,022,944} images with \textbf{2,219,146} annotated heads, over 20× larger than WIDERFace \cite{wider} while providing complete 3D annotations for every head. The generation process required 4,000 GPU hours but demonstrates the scalability: our pipeline can generate unlimited additional data as computational resources permit, offering an alternative to privacy-constrained real datasets.

\section{Method}

We present a unified architecture combining head detection and coarse 3D mesh estimation in a single forward pass, eliminating sequential processing pipelines. We extend object detection architectures to predict comprehensive 3D morphable model parameters alongside standard bounding box outputs.
Building upon YOLO-NAS \cite{yolonas}, we augment the real-time detector to regress FLAME \cite{FLAME} parameters directly from multi-scale feature maps. This design preserves single-stage detector efficiency while enabling comprehensive 3D modeling suitable for real-time applications. \method distills a traditional sequential pipeline (head detection → cropping → 3D regression) into a unified model that outperforms the original teacher through joint optimization and error accumulation elimination.
For each detected head, our model predicts a complete vector of 3DMM parameters disentangled into shape, expression, and pose components. This parametric representation enables recovery of head and face bounding boxes, 3D mesh vertices, pose rotations, and facial landmarks through differentiable rendering. Compared to existing methods with limited representations, our approach provides a unified foundation for diverse downstream tasks.
The architecture \cref{fig:model_arch}) employs six specialized prediction heads decoding FLAME's parametric structure: shape, expression, jaw pose, global head pose, translation, and scale. Each module operates on shared multi-scale features, enabling efficient computation while preserving semantic meaning of parameter groups.

\begin{figure*}[t!]
\includegraphics[width=0.99\textwidth]{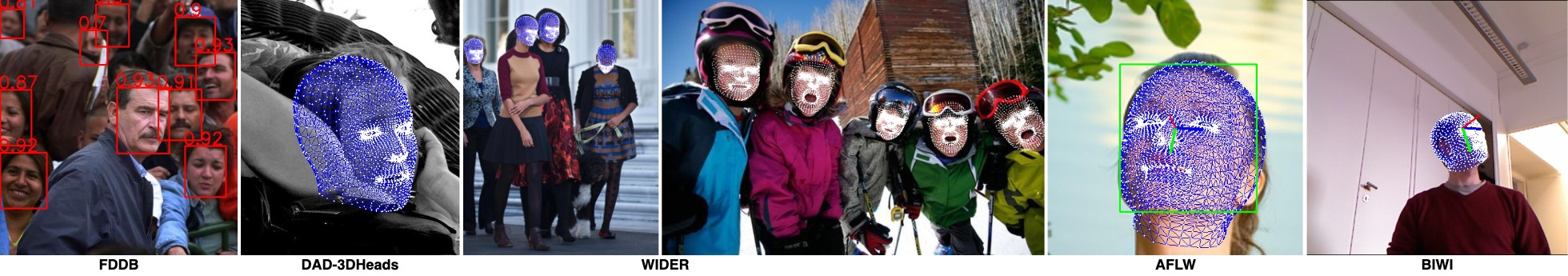}
\centering
\caption{\textbf{Versatility.} Our single model predicts many types of head annotations and works across all datasets.}
\label{fig:datasets}
\end{figure*}

\subsection{Unified Multi-Scale Architecture}
\method bridges the representation gap between coarse detection and parametric 3D modeling. The YOLO-NAS backbone extracts hierarchical features at multiple scales, enabling head detection across wide size ranges. Rather than treating detection and 3D estimation as separate problems, we jointly optimize both through shared feature representations.
Multi-scale design is crucial for multi-person scenarios where head sizes vary dramatically. Traditional crop-based methods struggle with scale variation due to normalized input assumptions. Our approach naturally handles scale diversity by leveraging multi-scale pyramids inherent in modern detection architectures.
Each detection head predicts traditional outputs (classification scores, bounding boxes) and 3D regression outputs (FLAME parameters). Joint prediction ensures 3D estimates are geometrically consistent with detected bounding boxes, eliminating misalignment issues in sequential approaches.

\subsection{Objective Functions} 
We design a multi-component loss providing supervision for all aspects of our unified representation. The loss combines standard detection objectives with novel 3D geometry losses, enabling end-to-end optimization.
Our loss consists of five components: two detection losses (Classification $L_C$ and Bounding Box Regression $L_{bbox}$) and three 3D losses (3D Vertices $L_{3D}$, Rotation $L_R$, and Reprojection $L_{proj}$). This combination ensures accurate detection and precise 3D estimation while maintaining computational efficiency.

\textbf{Reprojection Loss:} We measure the discrepancy between reprojected 3D vertices and ground truth 2D keypoint coordinates, providing direct supervision for the camera projection model:
\[ L_{\text{reproj}}(v_{p}, v_{gt}) = \frac{1}{N} \sum_{i=1}^{N} \left| v_p^i - v_{\text{gt}}^i \right| \]
where $N$ is the number of keypoints. This loss ensures that our 3D predictions project correctly onto the image plane, maintaining geometric consistency.

\textbf{3D Vertices Loss:} Following DAD-3D \cite{dad3d}, we compute L2 loss over normalized, canonical 3D head vertices. We set global rotation to zero to evaluate shape and expression predictions independently of pose:
\[ L_{3D}(v_{p}, v_{gt}) = \frac{1}{N} \sum_{i=1}^{N} \left| v_p^i|_{\mathbf{R}=\mathbf{0}} - v_{\text{gt}}^i \right|_2 \]
We subsample vertices by removing ears, eyeballs, and neck regions, then normalize both predicted and ground truth meshes to unit cubes. This canonical comparison enables robust shape supervision across different head sizes and orientations.

\textbf{Rotation Loss:} Rather than using standard Euler angle representations that suffer from discontinuities, we predict 6D rotation representations and apply geodesic distance loss on the resulting rotation matrices. This approach respects the SO(3) manifold geometry:
$$L_R(R_p, R_{gt}) = \cos^{-1} \left( \frac{\text{tr}(R_p R_{gt}^T) - 1}{2} \right) $$
The geodesic distance measures the shortest path between rotations, providing stable gradients.

\textbf{Detection Losses:} We employ focal loss $L_c$ \cite{focal_loss} for classification, addressing class imbalance in multi-person scenarios, and Complete IoU loss $L_{reg}$ \cite{zheng2020distance} for bounding box regression, incorporating center distance and aspect ratio for improved localization.

The final loss combines all components with learned weights:
$$L = \alpha_\mathrm{3D} L_\mathrm{3D} +  \alpha_\mathrm{R} L_\mathrm{R} + \alpha_\mathrm{reproj} L_\mathrm{reproj} + \alpha_\mathrm{c} L_\mathrm{c} + \alpha_\mathrm{reg} L_\mathrm{reg}$$
Our ablation studies (Table~\ref{t:model_abl}) demonstrate that each loss component contributes meaningfully to final performance, with the 3D losses enabling substantially better geometric understanding compared to detection-only baselines.

\subsection{Implementation Details} 

We implement our model in PyTorch, initializing the backbone with COCO \cite{mscoco} pre-trained weights to leverage general object detection knowledge. The differentiable FLAME layer remains fixed during training, with shape and expression parameters set to 300 and 100 dimensions respectively, providing sufficient representational capacity while maintaining computational efficiency.
Training requires 4 RTX A6000 GPUs with batch size 80, converging after 7 days. We resize all images to $640 \times 640$ pixels while preserving aspect ratios through padding, ensuring consistent head scales across the dataset. To bridge the synthetic-to-real domain gap, we apply extensive data augmentation including blur, noise, compression, and color manipulation, simulating real-world image degradations.
This implementation achieves real-time performance suitable for interactive applications while maintaining the accuracy necessary for downstream 3D modeling tasks, demonstrating the effectiveness of our unified architecture design.
\section{Experimental Evaluation}

\begin{table*}[t!]
\centering
\footnotesize
\caption{\textbf{3D Head Pose Estimation.} \method achieves state-of-the-art among end-to-end methods on AFLW \cite{aflw} and BIWI \cite{biwi}, and is competitive with non-end-to-end approaches.}\label{t:3dpose_aflw_biwi}
\begin{tabular}{@{}lllrrrr rrrr@{}}
\noalign{\smallskip}
\toprule %
 & & & \multicolumn{4}{c}{AFLW} & \multicolumn{4}{c}{BIWI} \\  \cmidrule{4-7} \cmidrule(l){8-11}
\textbf{Model} & \textbf{End to End} & \textbf{3DMM} & \textbf{MAE} ↓ & \textbf{Pitch} ↓ & \textbf{Roll} ↓ & \textbf{Yaw} ↓  & \textbf{MAE} ↓ & \textbf{Pitch} ↓ & \textbf{Roll} ↓ & \textbf{Yaw} ↓ \\\midrule %
6DRepNet \cite{repnet6d} & \xmark & \xmark & \textbf{3.61} & \textbf{4.58} & 2.98 & 3.27 & 3.78 & 5.32 & 2.78 & \textbf{3.23}\\
3DDFA-V2 \cite{3ddfav2} & \xmark & \cmark & 7.56 & 8.48 & 9.89 & 4.30 & 8.81 & 12.08 & 7.54 & 6.80\\
DAD-3DNet \cite{dad3d} & \xmark & \cmark & 3.66 & 4.76 & \textbf{3.15} & \textbf{3.08} & \textbf{3.98} & \textbf{5.24} & \textbf{2.92} & 3.79\\\midrule
RetinaFace \cite{retinaface} & \cmark & \xmark & 6.22 & 9.64 & 3.92 & 5.10  & 4.49 & 6.42  & 2.97 & 4.07\\
Img2Pose \cite{img2pose} & \cmark & \xmark & 3.91 & 5.03 & \textbf{3.28} & 3.43  & \textbf{3.79} & \textbf{3.55} & 3.24 & 4.57\\
\method & \cmark & \cmark & \textbf{3.76} & \textbf{4.91} & 3.37 & \textbf{3.00} & \textbf{3.79} & 5.24 & \textbf{2.65} & \textbf{3.47}\\
\bottomrule %
\noalign{\smallskip}
\end{tabular}
\end{table*}

We extensively evaluate both our synthetic dataset and unified architecture across multiple head-related tasks. Our evaluation demonstrates that: (1) models trained exclusively on synthetic data can achieve state-of-the-art performance on real benchmarks, (2) our dataset improves existing methods beyond our specific architecture, and (3) our unified representation enables superior performance across diverse downstream tasks.

\begin{table}
\centering
\footnotesize
\setlength{\tabcolsep}{1.5pt}
\caption{\textbf{Dataset Evaluation.} The model trained on \method shows better performance on 3D Head Alignment and Pose Estimation compared to training on traditional face and head datasets (FDDB \cite{fddb}, SCUT \cite{scut-head}, HH \cite{hollywood_heads} and WIDER \cite{wider}.}
\label{t:data_eval}
\begin{tabular}{@{}lccccccc@{}}
\noalign{\smallskip}
\toprule %
\multirow{2}{*}{\textbf{Dataset}}  &  \multicolumn{4}{c}{\textbf{DAD-3D}} & \textbf{AFLW} & \textbf{BIWI} & \textbf{FDDB}  \\ \cmidrule{2-5}\cmidrule(l){6-6} \cmidrule(l){7-7} \cmidrule(l){8-8} 
 & \textbf{NME}{$\downarrow$} & \textbf{\textit{Z}\textsubscript{5}} \textbf{Acc.}{$\uparrow$} & \textbf{CD}{$\downarrow$} & \textbf{PoseErr}{$\downarrow$} & \textbf{MAE}{$\downarrow$} & \textbf{MAE}{$\downarrow$} & \textbf{AP50}{$\uparrow$} \\\toprule %
FDDB & 6.22 & 0.84 & 7.21 & 0.51 & 8.42 & 5.23 & 97.1\\
SCUT & 5.57 & 0.84 & 6.23 & 0.43 & 8.91 & 5.44 & 84.1\\
HH & 3.91 & 0.91 & 5.02 & 0.29 & 8.12 & 4.92 & 89.2\\
WIDER  & 5.03 & 0.87 & 5.83 & 0.38 & 9.83 & 5.17 & \textbf{96.1}\\
\method & \textbf{2.92} & \textbf{0.93} & \textbf{4.00} & \textbf{0.18} & \textbf{3.67} & \textbf{3.58} & \textbf{96.1}\\
\bottomrule %
\noalign{\smallskip}
\end{tabular}
\end{table}
\begin{table}[h]
\centering
\footnotesize
\setlength{\tabcolsep}{2pt}
\caption{\textbf{Cross-Method Dataset Validation.} Training img2pose \cite{img2pose} on \method dataset improves head pose estimation and detection performance compared to WIDER.}
\label{t:cross_method_validation}
\begin{tabular}{@{}llccccc@{}}
\noalign{\smallskip}
\toprule %
\multirow{2}{*}{\textbf{Method}} & \multirow{2}{*}{\textbf{Dataset}} & \multicolumn{2}{c}{\textbf{AFLW}} & \multicolumn{2}{c}{\textbf{BIWI}} & \textbf{FDDB}  \\ 
\cmidrule{3-4}\cmidrule(l){5-6}\cmidrule(l){7-7}
 & & \textbf{MAE}{$\downarrow$} & \textbf{Yaw}{$\downarrow$} & \textbf{MAE}{$\downarrow$} & \textbf{Yaw}{$\downarrow$} & \textbf{AP50}{$\uparrow$} \\
\midrule
img2pose & WIDER & 3.91 & 3.43 & 3.79 & 4.57 & 96.1\\
img2pose & \method & \textbf{3.82} & \textbf{3.25} & \textbf{3.74} & \textbf{4.26} & \textbf{96.3}\\
\bottomrule %
\noalign{\smallskip}
\end{tabular}
\end{table}

\subsection{Dataset Evaluation}

We first validate our \method dataset by comparing our method's performance when trained on different datasets and by demonstrating cross-method improvements.

\textbf{Cross-Dataset Training:} \Cref{t:data_eval} shows our method trained on various traditional datasets compared to \method training. Training on individual traditional datasets (FDDB \cite{fddb}, SCUT-Head \cite{scut-head}, HollywoodHeads \cite{hollywood_heads}, WIDER \cite{wider}) consistently underperforms SH3D across all 3D (DAD-3D \cite{dad3d}) and pose estimation benchmarks (AFLW \cite{aflw}, BIWI \cite{biwi}). This demonstrates that our synthetic dataset provides superior training data on a large scale for 3D head modeling tasks, even when compared to real datasets.

\textbf{Cross-Method Validation:} \Cref{t:cross_method_validation} addresses the critical question of whether our dataset benefits existing methods beyond our architecture. We retrained img2pose on \method versus WIDER, showing consistent improvements across AFLW (3.91→3.82 MAE), BIWI (3.79→3.74 MAE), and FDDB (96.1→96.3 AP50). This proves our dataset's value is not limited to our specific method design.

\subsection{Head Pose Estimation}

We evaluate 3D head pose estimation accuracy on AFLW2000-3D~\cite{aflw} and BIWI~\cite{biwi} datasets. AFLW2000-3D comprises 2,000 subjects with 68 3D landmarks and pose annotations, while BIWI contains laboratory-recorded RGB-D data with head rotations up to ±75° yaw, ±60° pitch, and ±50° roll.

\textbf{Results:} \Cref{t:3dpose_aflw_biwi} shows \method outperforms end-to-end methods~\cite{retinaface, img2pose} and achieves comparable performance to crop-based 3DMM estimators that assume perfect head localization. This is significant because crop-based methods operate on pre-aligned, tightly cropped heads, while our method must first detect heads. Achieving superior performance on this more challenging problem validates both our synthetic data and unified architecture design.

\subsection{3D Head Alignment}

\begin{figure*}[htb]
\includegraphics[width=0.99\textwidth]{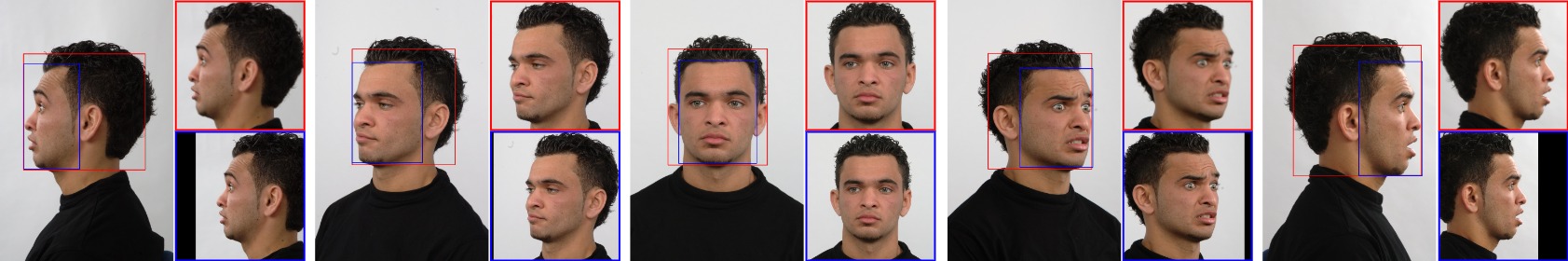}
\caption{\textbf{Head Alignment.} \method introduce more consistent alignment across various poses by ensuring the center of the head in 3D is reprojected to the center of the aligned image. \textcolor{red}{\method}, \textcolor{blue}{RetinaFace \cite{retinaface}.}}
\label{fig:head_align}
\end{figure*}

We utilize the DAD-3DHeads Benchmark \cite{dad3d} to evaluate 3D dense head alignment and robustness to extreme poses. The benchmark consists of 2,746 images with FLAME topology meshes, measuring pose fitting and shape matching under challenging conditions including extreme poses, illumination, and occlusions. For multi-head images, we extract the head with highest ground truth bounding box overlap.

\textbf{Results:} \Cref{t:dad} shows \method significantly outperforms RingNet \cite{RingNet} and 3DDFA-v2 \cite{3ddfav2}, despite these methods being trained on real images and optimized for tight crops. We achieve comparable results to DAD-3D \cite{dad3d} even though it was trained on the same distribution as this benchmark. Our FLAME-based representation provides geometric consistency with the skull center consistently projecting to image center regardless of orientation (\cref{fig:head_align}), essential for downstream 3D modeling tasks.

\begin{table}
\centering
\footnotesize
\setlength{\tabcolsep}{2.5pt}
\caption{\textbf{DAD3D-Heads.} \method achieves superior performance to RingNet and 3DDFA-V2, while being only slightly inferior to the methods trained on the benchmark. E2E: end to end.}
\label{t:dad}
\begin{tabular}{@{}llcccc@{}}
\noalign{\smallskip}
\toprule %
\textbf{Model} & \textbf{E2E} & \textbf{NME}{$\downarrow$} & \textbf{\textit{Z}\textsubscript{5}} \textbf{Acc.}{$\uparrow$} & \textbf{Chamf.~Dist.}{$\downarrow$} & \textbf{Pose Err.}{$\downarrow$} \\\toprule %
3DDFA-V2 \cite{3ddfav2} & \xmark & 3.580 & - & 6.170 & 0.527\\
RingNet \cite{RingNet} & \xmark & 8.757 & 0.880 & 5.166 & 0.438\\
DAD-3DNet \cite{dad3d} & \xmark  & 2.302 & 0.954 & 3.178 & 0.138\\
\method & \cmark & 2.917 & 0.933 & 4.002 & 0.179\\
\bottomrule %
\noalign{\smallskip}
\end{tabular}
\end{table}

\begin{table}[t]
\centering
\setlength{\tabcolsep}{3.5pt}
\footnotesize
\caption{\textbf{Face Detection.} The model trained on \method data is not optimized for tiny faces detection but with additional finetuning it shows comparable results to state-of-the-art detectors while recovering more complete representation.}
\label{t:face_det}
\begin{tabular}{@{}llcccc@{}}
\noalign{\smallskip}
\toprule %
\multirow{2}{*}{\textbf{Model}} & \multirow{2}{*}{\textbf{Dataset}}  & \textbf{FDDB} &  \multicolumn{3}{c}{\textbf{WIDER Val (AP50)}}  \\
 & & \textbf{AP50}{$\uparrow$} & {\textbf{Easy}{$\uparrow$}} & {\textbf{Med.}{$\uparrow$}} & {\textbf{Hard}{$\uparrow$}} \\\toprule %
RetinaFace \cite{retinaface} & WIDER & 96.2 & \textbf{94.6} & \textbf{93.0} & \textbf{80.4} \\
img2pose \cite{img2pose} & WIDER & 96.1 & 86.5 & 82.9 & 61.3 \\ \midrule
\method & \method & 96.1 & 56.3 & 51.0 & 29.2\\
\method & \method + WIDER & \textbf{96.6} & 92.6 & 88.9 & 70.3\\
\bottomrule %
\noalign{\smallskip}
\end{tabular}
\end{table}







\begin{table}
\caption{\textbf{Ablation.} Removing any loss reduces the performance on all three datasets.}\label{t:model_abl}
\begin{tabular}{@{}lccccc@{}}
\toprule
\multirow{2}{*}{\textbf{Model}} & \multicolumn{3}{c}{\textbf{\method Val}} & \textbf{AFLW} & \textbf{FDDB} \\
& \textbf{NME}{$\downarrow$}  &  \textbf{FR}{$\downarrow$}  &  \textbf{MAE}{$\downarrow$}  &  \textbf{MAE}{$\downarrow$} & \textbf{AP50}{$\uparrow$} \\\toprule
\method & \textbf{1.975} & 0.122 & \textbf{2.470} & \textbf{5.03} &\textbf{88.5} \\ \midrule
w/o $L_R$ & 2.052 & \textbf{0.113} & 2.863 & 6.27 & 87.9 \\
w/o $L_{3D}$ & 2.050 & 0.128 & 2.600 & 5.13 & 87.9 \\
\bottomrule
\end{tabular}
\end{table}


\subsection{Face Detection}

The last step is to validate the model’s ability to detect heads and faces under different conditions. While our model predicts head bounding boxes, we can derive face boxes from FLAME vertex projections, enabling evaluation on established face detection benchmarks \cite{fddb, wider}.

For face box computation, we calculate minimum bounding rectangles around facial vertex subsets and filter detections where $|\text{yaw}| > \frac{\pi}{2}$ (profile views where faces are not visible). Importantly, unlike methods explicitly optimized for face detection, we do not target extremely small faces (few pixels wide) since 3DMM parameter estimation becomes highly ambiguous at such scales. There is insufficient visual information for reliable 3D reconstruction.

\textbf{Results:} \cref{t:face_det} shows that \method achieves comparable performance on FDDB \cite{fddb} to specialized face detectors. This is remarkable considering \method recovers a more complete head representation and is not optimized for face detection. The performance validates that our unified approach does not sacrifice detection quality while gaining substantial 3D modeling capabilities.

\subsection{Ablation Study}

 We analyze the contribution of our novel loss components by systematically removing them during training \cref{t:model_abl}. We ablate the 3D vertices loss ($L_{3D}$) and rotation loss ($L_R$) as these represent the 3D supervision contributions, while the standard detection losses ($L_{reproj}$, $L_c$, $L_{reg}$) are essential for basic functionality and cannot be removed. Each ablated component proves essential: the 3D vertices loss enables better shape reconstruction through direct geometric supervision, while the geodesic rotation loss significantly reduces pose estimation errors via manifold-aware optimization. Ablation experiments use our manually annotated subset for computational efficiency.
\section{Conclusion}
We have presented \method, a unified approach that bridges head detection and coarse 3D mesh estimation in a single forward pass, thereby eliminating fragmented (pre-)processing pipelines.
Training this model was only possible using our new synthetic dataset, since it is the first to contain multi-task labels. Our large-scale synthetic dataset, comprising over 1 million multi-person images, captures unconstrained scenarios while addressing privacy concerns that have plagued major real-world datasets. 
Importantly, we demonstrate that models trained exclusively on synthetic data achieve state-of-the-art performance on real-world benchmarks, often outperforming methods trained on real data despite solving the more challenging problem of joint detection and parametric 3D modeling.
Our coarse 3D estimates provide robust initialization for detailed reconstruction methods, bridging the gap between detection and high-fidelity modeling. 
The dataset, code, and models will be made available, supporting further research in unified head modeling and the broader potential of synthetic data generation.

\textbf{Acknowledgements.} We would like to thank Tetiana Martyniuk and Iro Laina for paper proofreading and valuable feedback. O.K. is supported by a Google unrestricted gift. We also thank the Armed Forces of Ukraine for providing security to complete this work.

{
    \small
    \bibliographystyle{ieeenat_fullname}
    \bibliography{paper}
}
\clearpage

\appendix
\section{Data Quality} 

\begin{table}[h]
\centering
\caption{\textbf{Generators.} We experiment with different image generators \cite{ldm, sdxl} and control mechanisms \cite{controlnet, t2i}. Our method is not specific to any combination and improves with better generators.}\label{t:gen_abl}
\begin{tabular}{@{}llcc@{}}
\noalign{\smallskip}
\toprule %
\textbf{Model} & \textbf{Condition}& \textbf{FID}{$\downarrow$} & {\textbf{IS}{$\uparrow$}} \\\toprule %
SD 1.5 & ControlNet & 7.86 & 13.59 \\
SDXL & ControlNet & 4.18 & \textbf{15.09} \\
SDXL & T2I  &\textbf{3.22} & 14.37 \\
\bottomrule %
\noalign{\smallskip}
\end{tabular}
\end{table}

We evaluate the impact of different diffusion models and conditioning mechanisms by generating 20,000 images per configuration. Using FID \cite{fid} and Inception Score \cite{inception_score}, we measure scene diversity and realism, evaluating full images rather than face crops focusing on scene composition and multi-person layout quality, crucial for robust real-world performance.

\cref{t:gen_abl} demonstrates that advanced diffusion models \cite{sdxl} significantly outperform earlier versions \cite{ldm} in generating high-quality multi-person scenes. While ControlNet \cite{controlnet} and T2I-Adapter \cite{t2i} achieve comparable metrics, ControlNet frequently produces anatomical deformations, leading us to choose T2I-Adapter for more reliable scene generation.

\section{Re-Identification on Synthetic Data}

\begin{figure*}[htb]
\includegraphics[width=0.99\textwidth]{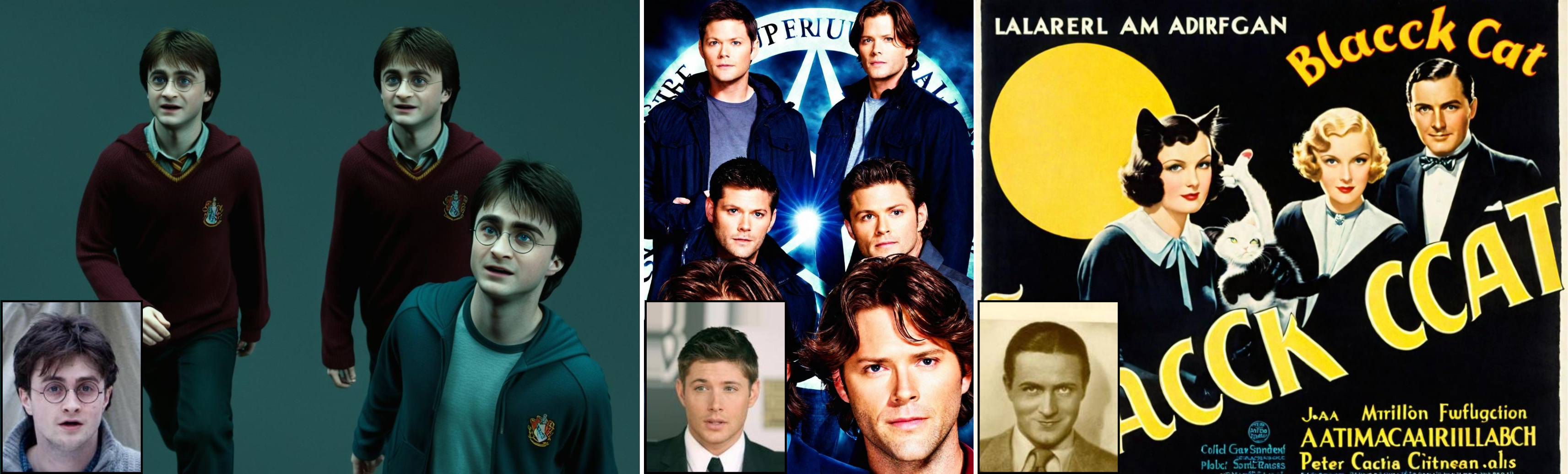}
\centering
\caption{\textbf{Re-ID on Celeb-A.} We automatically detect and removed samples where face in the generated image is matched with one of the faces from Celeb-A dataset \cite{celeba}.}
\label{fig:reid_celeb}
\end{figure*}

\begin{figure*}[htb]
\includegraphics[width=0.33\textwidth]{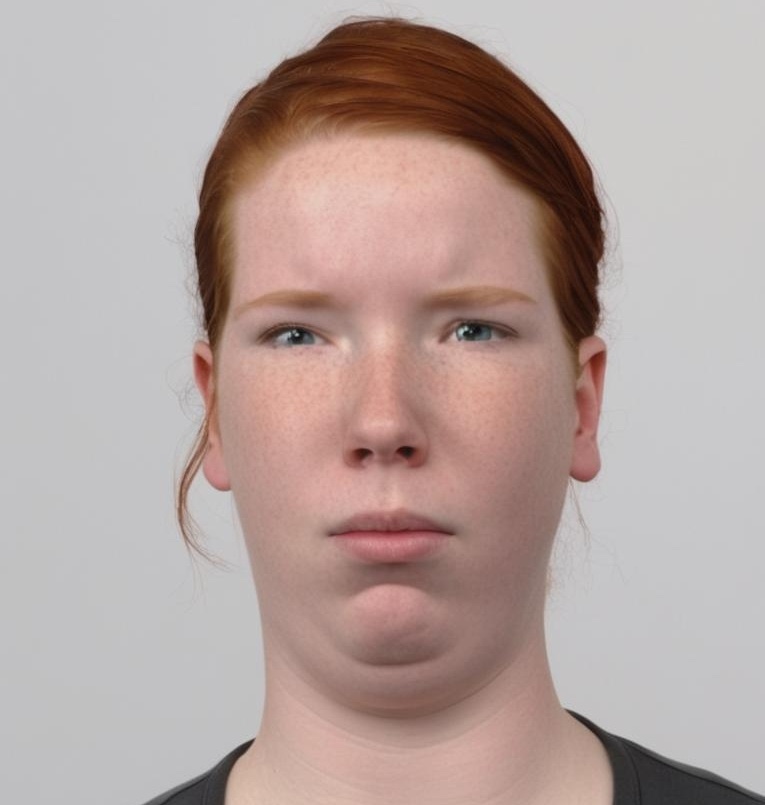}
\includegraphics[width=0.33\textwidth]{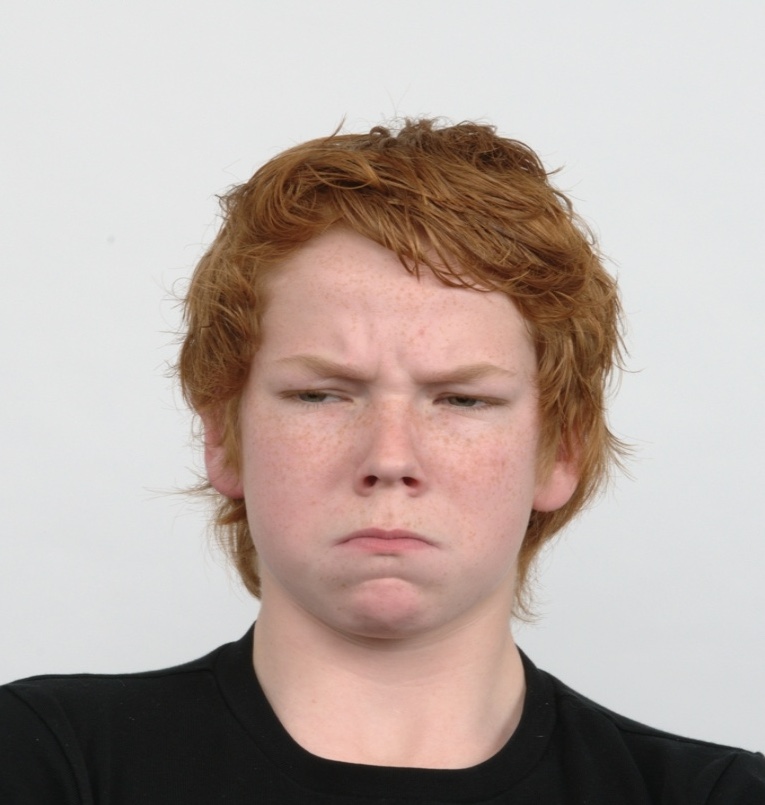}
\includegraphics[width=0.33\textwidth]{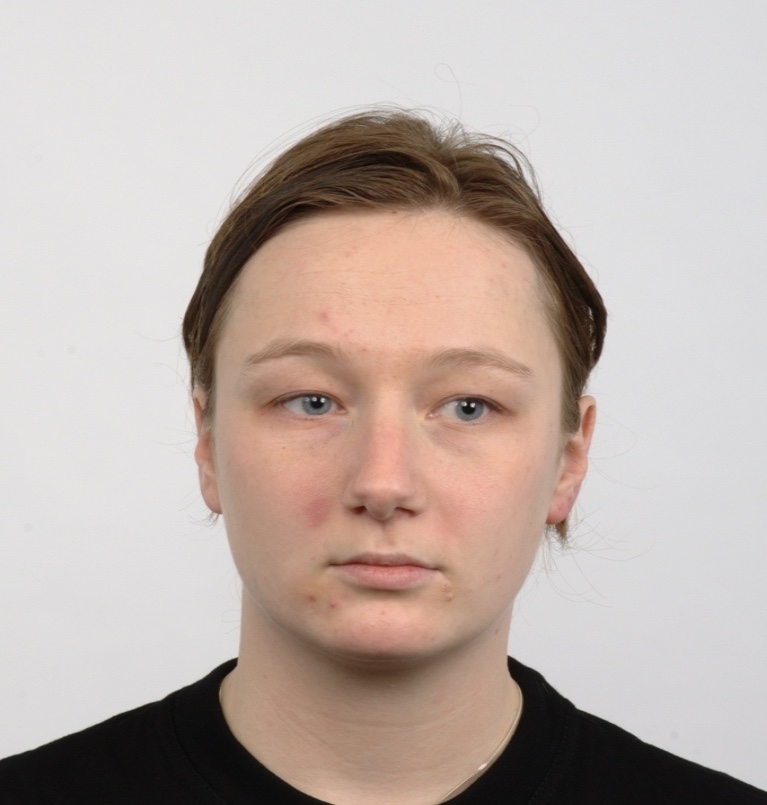}
\caption{\textbf{Preserving Identity.} With subject neutral prompts diffusion Model blends feature of different people from the set it was trained on.}
\label{fig:rafd_reid}
\end{figure*}

\Cref{fig:reid_celeb} provides visual examples of generated samples that were matched with public figures from Celeb-A \cite{celeba} dataset. Even though the data generation include the prompt anonymization, some parts of the prompts, such as the movie name still might encode identity of famous people. Identifying and removing such samples improve the privacy aspect.
\cref{fig:rafd_reid} shows an example of an image generated by diffusion model that was finetuned on RaFD \cite{rafd} dataset, and 2 nearest neighbors from the dataset it was trained on. Even though the model learns features from the original images, in most cases it does combine features from multiple people and fails to recover finer details that encode the identity.

\section{Model Details}
\label{sec:arch_details}

The architecture is inspired by object detection methods like CenterNet \cite{centernet} or YOLO-NAS-Pose \cite{yolonas}, enabling efficient end-to-end learning.
The model is based on the YOLO-NAS \cite{yolonas} architecture for object detection. YOLO-NAS use a neural architecture search engine to enhance the YOLO family of models by optimzing the sizes and structures of stages, block types, the number of blocks, and the number of channels in each stage. We employ the YOLO-NAS-L backbone, though the model is agnostic to the choice of encoder. The ``neck'' is used to fuse the features generated by the backbone. The visual features from the encoder maps neck are fused by the Spatial Pyramid Pooling \cite{spp} module at different scales and processed by Feature Pyramid Network \cite{fpn} to generate features at different semantic levels. Similarly to other YOLO models, we adopt an anchor-based multi-scale detection scheme. Path Aggregation Network (PAN) \cite{pan} transfers positioning features bottom-up. We combine them with the features from FPN to obtain a better feature fusion effect and then directly use the multi-scale fusion feature maps in the PAN for detection. Thus, the detection heads predict bounding boxes and 3DMM parameters on different scales, ensuring high accuracy on different object sizes.
This approach allows us to optimize FLAME parameters only on positive anchors, improving training efficiency. It also offers flexibility for applications that may not require full 3D meshes, allowing the extraction of bounding boxes in real time without computing full 3D mesh.
The detection head in the YOLO-NAS model predicts the offset of the bounding box position
and the scaling of the height and width, as well as the confidence of the prediction. We extend the detection head to also predict the 3DMM parameters by introducing six separate 3D parameter prediction modules, each consisting of two RepVGG blocks \cite{repvgg} and a final $1\times 1$ convolution that predicts the final set of parameters. Each RepVGG block consists of three branches: a $3\times 3$ Convolution followed by BatchNorm \cite{batchnorm}, a branch of a 1x1 Convolution with bias and a residual branch. Predicting different 3DDM parameters components separately achieves an extra level of disentanglement.

The weights of the loss components are set to \( \alpha_1 = 50, \alpha_2 = 1, \alpha_3 = 1, \alpha_4 = 0.5, \alpha_5 = 2.5 \) in the final version.

 The framework is agnostic to the choice of backbone and can be adapted to use larger transformer models \cite{swin}. Yet, we observe the string performance on a smaller fully convolutional models and stick to this design to allow for real-time multi head mesh recovery. Furthermore the versatility of backbones allow to train even smaller models suitable for wide range of tasks and applications.
\begin{table}
\centering
\footnotesize
\caption{\textbf{Model Architecture Comparison.} Analysis of different \method variants in terms of parameters, computational cost, and speed.}
\label{t:flops}
\begin{tabular}{@{}l rrr@{}}
\noalign{\smallskip}
\toprule
\textbf{Model} & \textbf{Total Parameters} & \textbf{FLOPs (B)} & \textbf{FPS} \\\midrule
\method-L & 50,442,706 & 83.51 & 60.13 \\
\method-M & 32,378,236 & 52.01 & 69.38 \\
\method-S & 17,004,954 & 22.92 & 72.34 \\
\bottomrule
\noalign{\smallskip}
\end{tabular}
\end{table}

\section{Head Pose Estimation}

The full results on AFLW and BIWI datasets are presented in \cref{t:3dpose_aflw_full}, \cref{t:3dpose_biwi_full}. \method outperforms most of method optimized solely for 3D Head Pose Estimation even though it does not operate on tight head crops.
\begin{table*}
\centering
\footnotesize
\caption{AFLW}\label{t:3dpose_aflw_full}
\begin{tabular}{@{}lllrrrr@{}}
\noalign{\smallskip}
\toprule %
\textbf{Model} & \textbf{End to End} & \textbf{3DMM} & \textbf{MAE} ↓ & \textbf{Pitch MAE} ↓ & \textbf{Roll MAE} ↓ & \textbf{Yaw MAE} ↓ \\\toprule %
Dlib \cite{dlib} & \xmark & \xmark & 13.29 & 12.60 & 9.00 & 18.27\\
HopeNet \cite{hopenet} & \xmark & \xmark & 6.16 & 6.56 & 5.44 & 6.47\\
6DRepNet \cite{repnet6d} & \xmark & \xmark & 3.61 & 4.58 & 2.98 & 3.27\\
\bottomrule %
RingNet \cite{RingNet} & \xmark & \cmark & 8.27 & 4.39 & 13.51 & 6.92\\
3DDFA-V2 \cite{3ddfav2} & \xmark & \cmark & 7.56 & 8.48 & 9.89 & 4.30\\
3DDFA \cite{3ddfa_cleardusk} & \xmark & \cmark & 7.39 & 8.53 & 7.39 & 5.40\\
DAD-3DHeads \cite{dad3d} & \xmark & \cmark & 3.66 & 4.76 & 3.15 & 3.08\\
SynergyNet \cite{synergynet} & \xmark & \cmark & \textbf{3.35} & \textbf{4.09} & \textbf{2.55} & 3.42\\
\bottomrule %
RetinaFace \cite{retinaface} & \cmark & \xmark & 6.22 & 9.64 & 3.92 & 5.10\\
Img2Pose \cite{img2pose} & \cmark & \xmark & 3.91 & 5.03 & \textbf{3.28} & 3.43\\
\method & \cmark & \cmark & \textbf{3.76} & \textbf{4.91} & 3.37 & \textbf{3.00}\\
\bottomrule %
\noalign{\smallskip}
\end{tabular}
\end{table*}
\begin{table*}
\centering
\footnotesize
\caption{BIWI}
\begin{tabular}{@{}lllrrrr@{}}
\noalign{\smallskip}
\toprule %
\textbf{Model} & \textbf{End to End} & \textbf{3DMM} & \textbf{MAE} ↓ & \textbf{Pitch MAE} ↓ & \textbf{Roll MAE} ↓ & \textbf{Yaw MAE} ↓ \\\toprule %
Dlib (68 points) \cite{dlib} & \xmark & \xmark & 12.25 & 13.80 & 6.19 & 16.76\\
HopeNet \cite{hopenet} & \xmark & \xmark & 4.90 & 6.61 & 3.27 & 4.81\\
WHENet \cite{whenet} & \xmark & \xmark & 3.81 & 4.39 & 3.06 & 3.99 \\
6DRepNet \cite{repnet6d} & \xmark & \xmark & 3.78 & 5.32 & 2.78 & \textbf{3.23}\\
MNN \cite{mnn} & \xmark & \xmark & \textbf{3.66} & 4.61 & \textbf{2.39} & 3.98 \\
\bottomrule
3DDFA \cite{3ddfa_cleardusk} & \xmark & \cmark & 19.07 & 12.25 & 8.78 & 36.18\\
3DDFA-V2 \cite{3ddfav2} & \xmark & \cmark & 8.81 & 12.08 & 7.54 & 6.80 \\
RingNet \cite{RingNet} & \xmark & \cmark & 7.34 & 5.37 & 7.82 & 8.82 \\
DAD-3DNet \cite{dad3d} & \xmark & \cmark & 3.98 & 5.24 & 2.92 & 3.79\\
\bottomrule
RetinaFace \cite{retinaface} & \cmark & \xmark & 4.49 & 6.42  & 2.97 & 4.07 \\
Img2Pose \cite{img2pose} & \cmark & \xmark & \textbf{3.79} & \textbf{3.55} & 3.24 & 4.57\\
\method & \cmark & \cmark & \textbf{3.79} & 5.24 & \textbf{2.65} & \textbf{3.47}\\\bottomrule %
\noalign{\smallskip}
\end{tabular}
\label{t:3dpose_biwi_full}
\end{table*}

\section{3D Face Reconstruction}

\method is more coarse than 3D reconstruction methods so detailed 3D face reconstruction is not the goal of our method and a standalone task in itself. Nonetheless, we evaluate the model on the Feng et al. benchmark \cite{feng}. The model achieves comparable results to other coarse 3D face reconstruction methods \cite{3ddfav2, RingNet, dad3d} despite not being optimized for shape and expression disentanglement. This performance is surprising since our method predicts the 3D face from a full image, not a tight crop as is typical for this task. While our method aims at solving a different task, it achieves good performance. Moreover, the \method dataset and model can be complementary to detailed face reconstruction methods such as DECA \cite{deca}, MICA \cite{mica} or EMOCA \cite{emoca}, which often need to be initialized with a crop or initial coarse face shape.

\begin{figure*}[htb]
\includegraphics[width=0.99\textwidth]{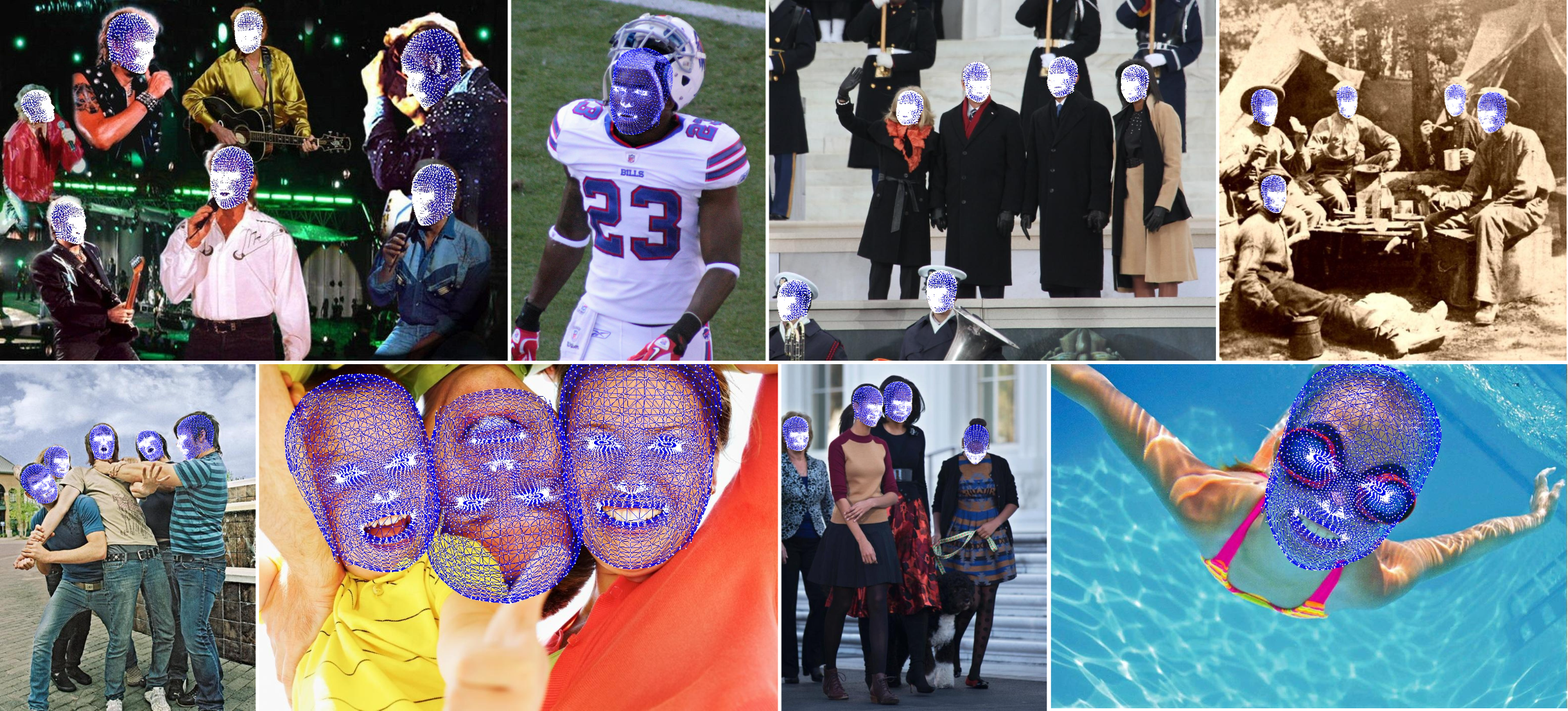}
\caption{\textbf{Qualitative Evaluation.} \method is able to accurately recover 3D head models on various complex scenes from WIDER Face \cite{wider} dataset.}
\label{fig:data_example}
\end{figure*}

\begin{figure*}[htb]
\includegraphics[width=0.99\textwidth]{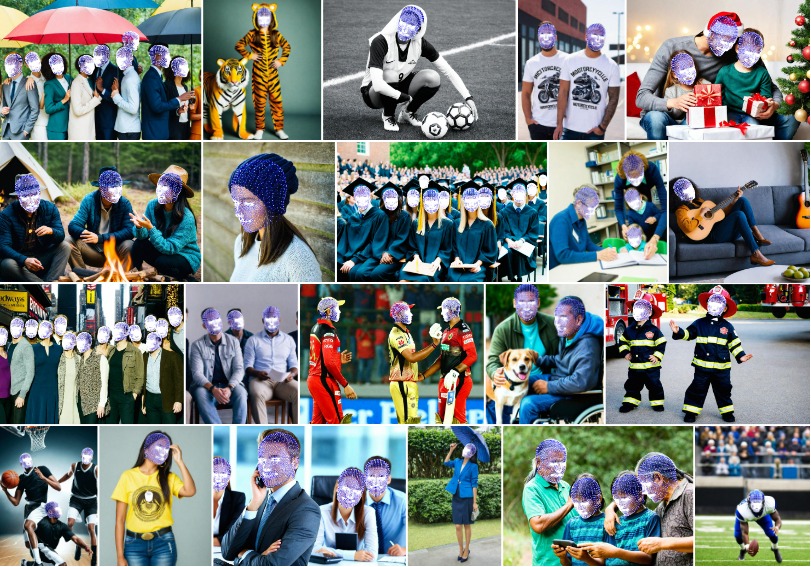}
\caption{\textbf{Dataset Examples.} The synthetic data generation pipeline generates complex realistic real world scenes with multiple objects, covering wide range of poses and backgrounds while reducing age, gender and ethnical biases present in small real world datasets.}
\label{fig:large_data_sample}
\end{figure*}

\begin{table}[h]
\footnotesize
\begin{adjustbox}{width=0.49\textwidth}
\begin{tabular}{@{}lccccccc@{}}
\noalign{\smallskip}
\toprule %
    \multirow{2}{*}{\textbf{Model}}& \multirow{2}{*}{\textbf{3DRMSE}↓} & 
    \multicolumn{2}{c}{\textbf{Median(mm)}↓} &
    \multicolumn{2}{c}{\textbf{Mean(mm)}↓} & 
    \multicolumn{2}{c}{\textbf{Std(mm)}↓} \\
             &       &   HQ  &   LQ  &   HQ  &   LQ  &   HQ  &   LQ  \\\toprule %
    3DDFA-V2\cite{3ddfav2} &   2.998 &   1.500  &   1.779  &   1.942   &  2.350    &   1.704  & 2.149  \\
    RingNet\cite{RingNet}  &   2.809   &   1.698   &   1.634   &   2.161  &   2.113    &  1.832   & 1.831  \\
    DAD-3DNet \cite{dad3d} &   2.749   &   1.558   &   1.624   &   1.940   &   2.082   &   1.581   &   1.795   \\
    \method &   2.996   &   1.622   &   1.801  &   2.079   &   2.353   &   1.801   &   2.054   \\\bottomrule
\noalign{\smallskip}
\end{tabular}
\end{adjustbox}
\caption{\textbf{Feng et al.\cite{feng}}}
\label{t:3d_recon}
\end{table}

\section{Full Body Mesh Recovery}

Our approach proves the viability of using synthetic data from diffusion models for body modeling, paving the way for future fully synthetic all-in-one methods. \method is the first step towards this goal. While full-body reconstruction methods \cite{pixie, pymaf}, achieved significant progress and attention, they often still rely on upstream face predictors and lack robustness in edge cases. To validate it we evaluate PIXIE's \cite{pixie} performance on the BIWI \cite{biwi} dataset for 3D Head Pose.

\begin{table}
\centering
\footnotesize
\caption{\textbf{Comparison with Full Body Reconstruction \cite{pixie}.} \method achieves superior performance to full body recovery method \cite{pixie} on BIWI 3D Head Pose Estimation. This shows that all in one methods still fail on challenging head understanding benchmarks.}
\label{t:full_body}
\begin{tabular}{@{}l rrrr@{}}
\noalign{\smallskip}
\toprule
\textbf{Model} & \textbf{MAE}↓ & \textbf{Pitch MAE}↓ & \textbf{Roll MAE}↓ & \textbf{Yaw MAE}↓ \\\midrule
PIXIE \cite{pixie} & 10.97 & 16.80 & 6.19 & 9.93 \\
\method & \textbf{3.79} & \textbf{5.24} & \textbf{2.65} & \textbf{3.47} \\
\bottomrule
\noalign{\smallskip}
\end{tabular}
\end{table}

\section{Controllable Generation}

\begin{figure*}[htb]
\includegraphics[width=0.99\textwidth]{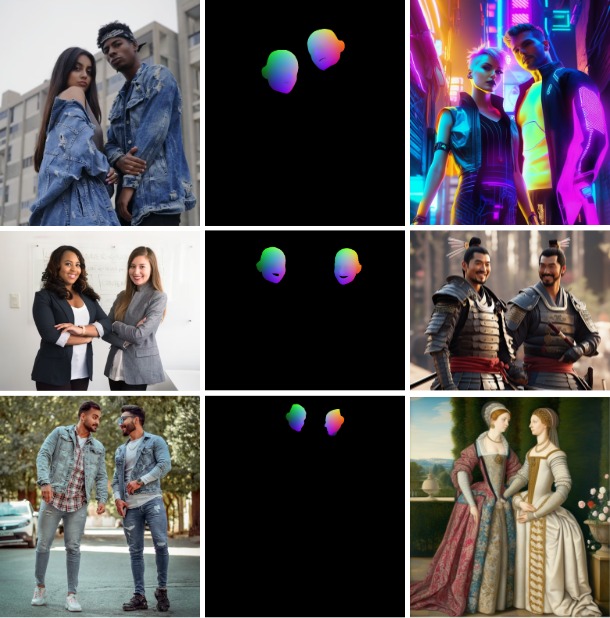}
\caption{\textbf{ControlNet with \method.} The 3D condition provides a strong degree of control for the generative model, preserving shape, pose and expression of the input image.}
\label{fig:pncc}
\end{figure*}

Conditioning the image generation on full head mesh helps to preserve the head shape and expression which is crucial for many AR applications. We trained the ControlNet \cite{controlnet} for SDXL \cite{sdxl} model that is conditioned on meshes recovered by our model. The meshes are rendered by mapping to RGB space with Projected Normalized Coordinate Code (PNCC) \cite{face_align}, with the 3D coordinate of each vertex of the normalized head mesh encoded as RGB  (NCCx = R, NCCy = G, NCCz = B). The heads on the generated images preserve the pose, expression and shape of the original photo \cref{fig:pncc}.

\section{Qualitative Results}
Additional data samples from \method dataset are presented in \cref{fig:large_data_sample}.

\begin{figure*}[htb]
\includegraphics[width=0.99\textwidth]{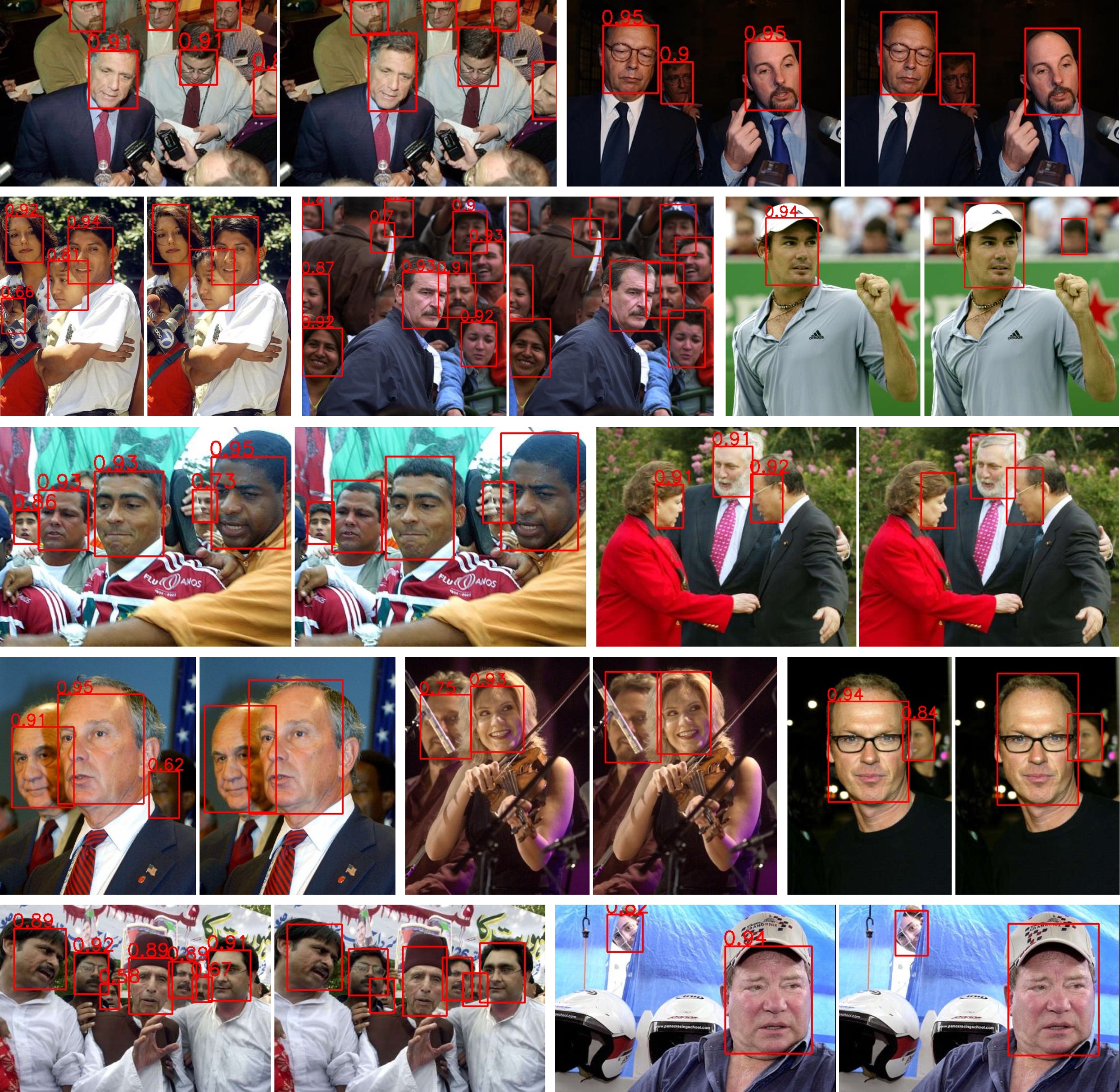}
\caption{\textbf{Qualitative Evaluation on FDDB.}}
\label{fig:fddb_qual}
\end{figure*}

\begin{figure*}[htb]
\centering
\includegraphics[width=0.23\textwidth]{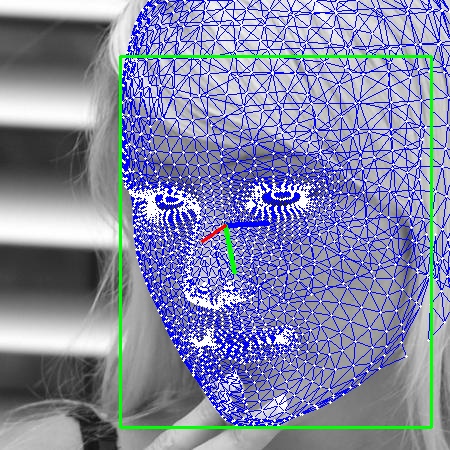}
\includegraphics[width=0.23\textwidth]{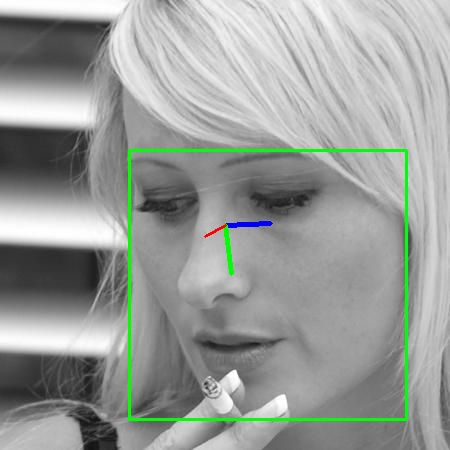}
\includegraphics[width=0.23\textwidth]{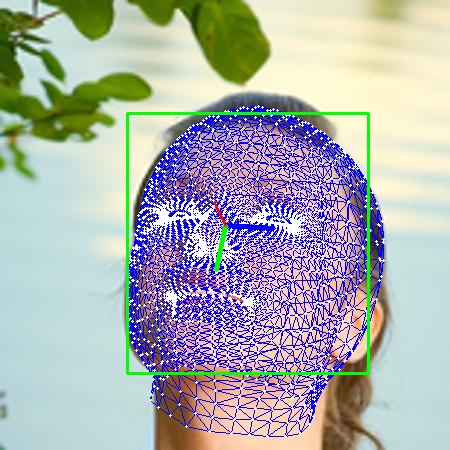}
\includegraphics[width=0.23\textwidth]{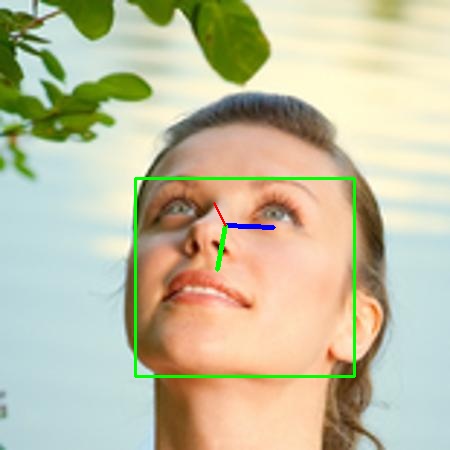}
\includegraphics[width=0.23\textwidth]{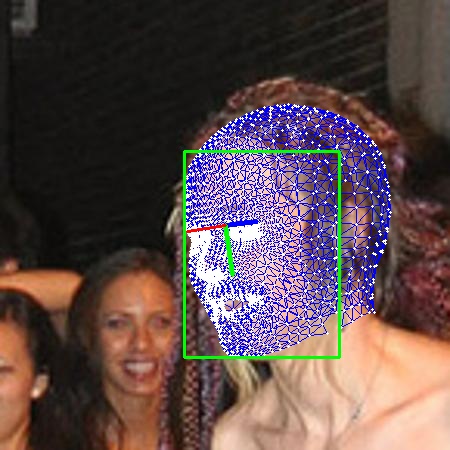}
\includegraphics[width=0.23\textwidth]{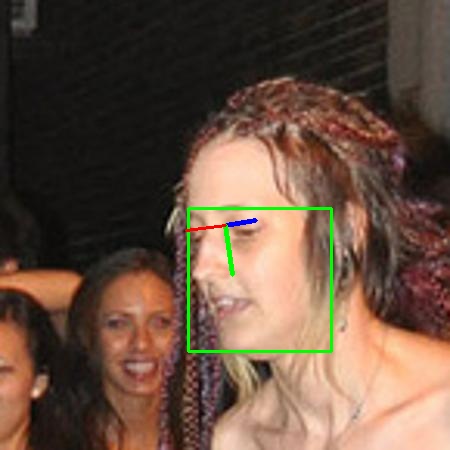}
\includegraphics[width=0.23\textwidth]{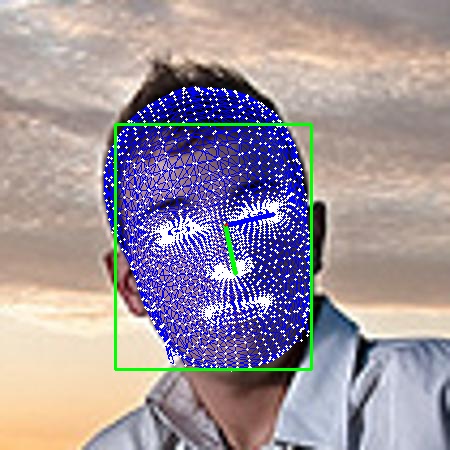}
\includegraphics[width=0.23\textwidth]{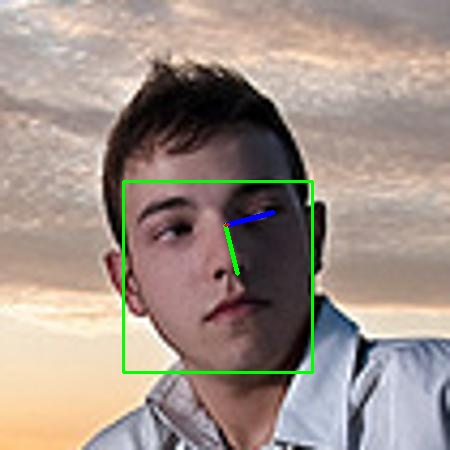}
\includegraphics[width=0.23\textwidth]{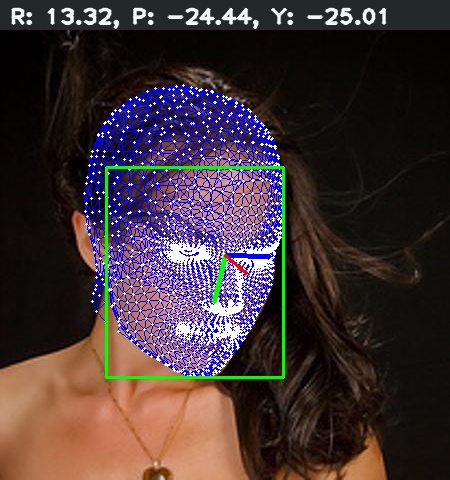}
\includegraphics[width=0.23\textwidth]{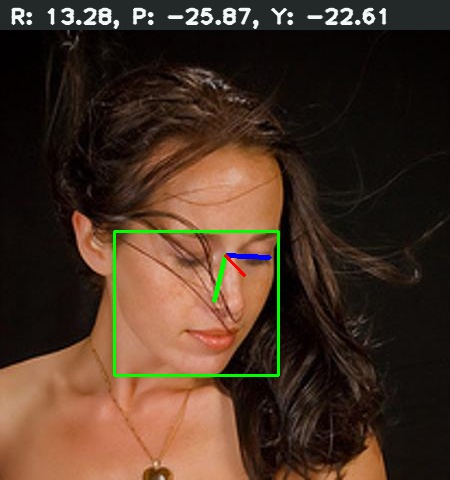}
\includegraphics[width=0.23\textwidth]{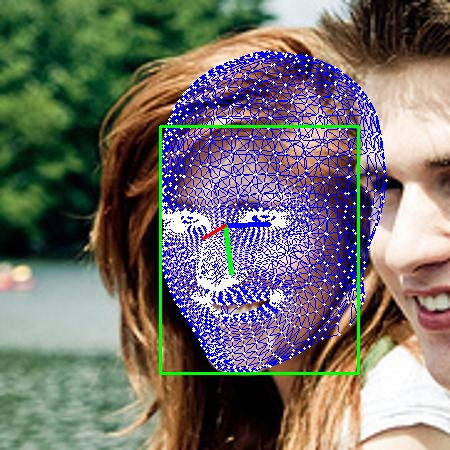}
\includegraphics[width=0.23\textwidth]{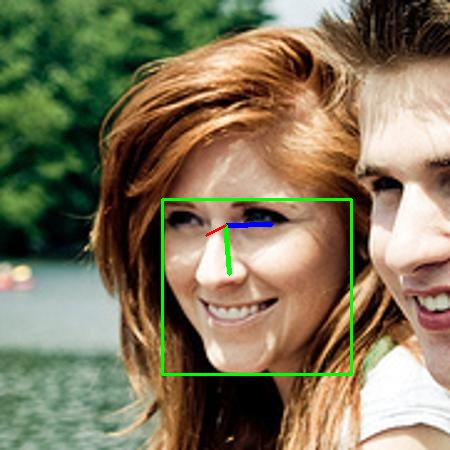}
\includegraphics[width=0.23\textwidth]{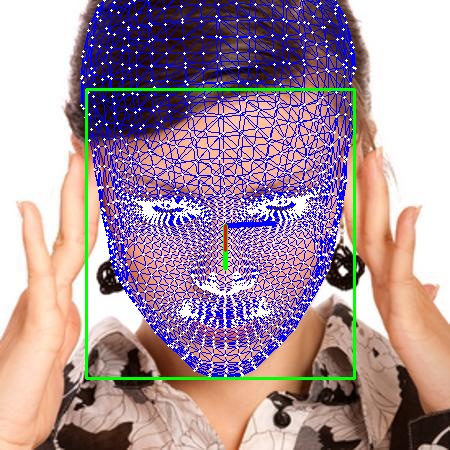}
\includegraphics[width=0.23\textwidth]{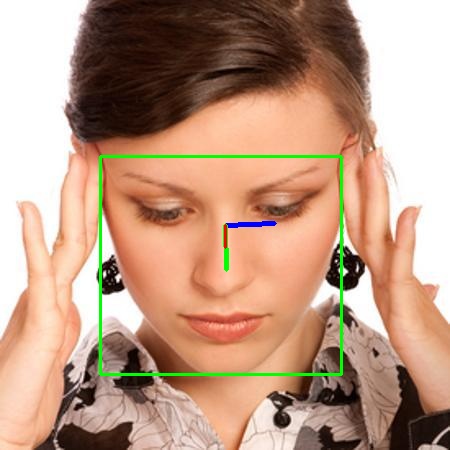}
\includegraphics[width=0.23\textwidth]{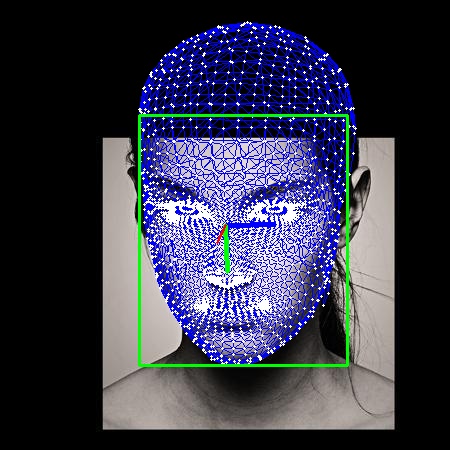}
\includegraphics[width=0.23\textwidth]{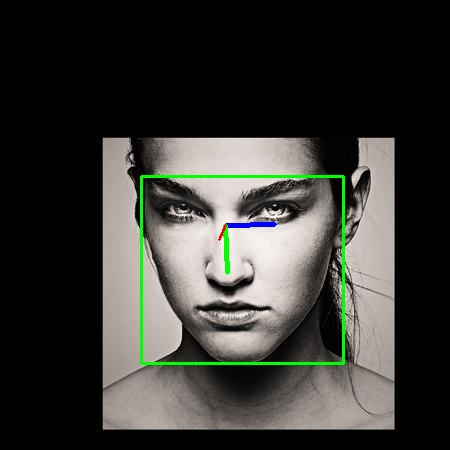}
\includegraphics[width=0.23\textwidth]{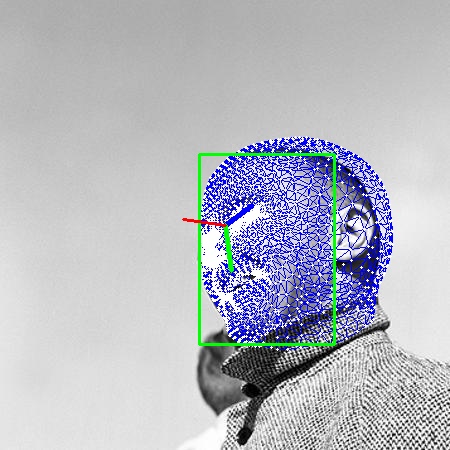}
\includegraphics[width=0.23\textwidth]{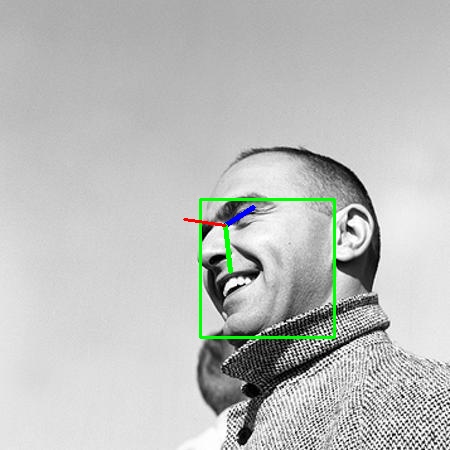}
\includegraphics[width=0.23\textwidth]{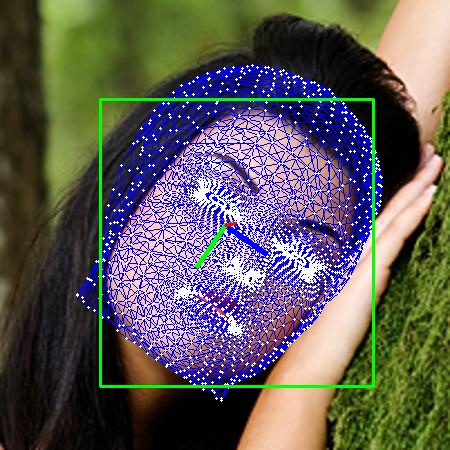}
\includegraphics[width=0.23\textwidth]{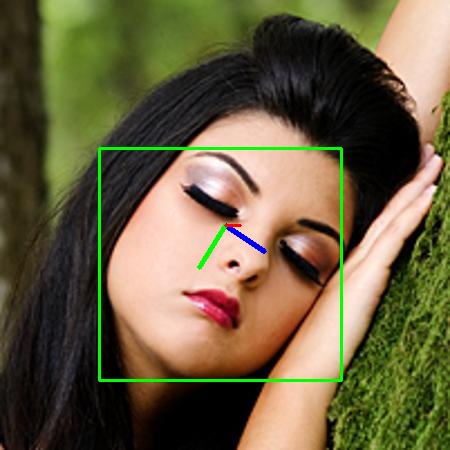}
\includegraphics[width=0.23\textwidth]{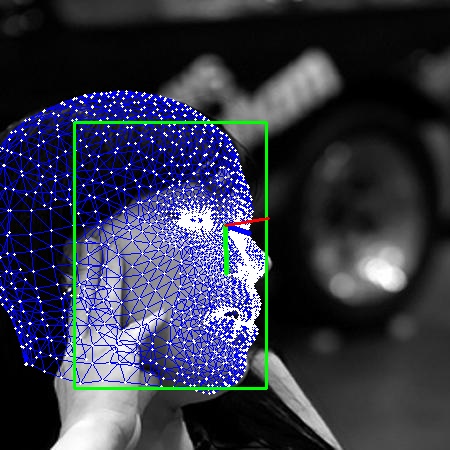}
\includegraphics[width=0.23\textwidth]{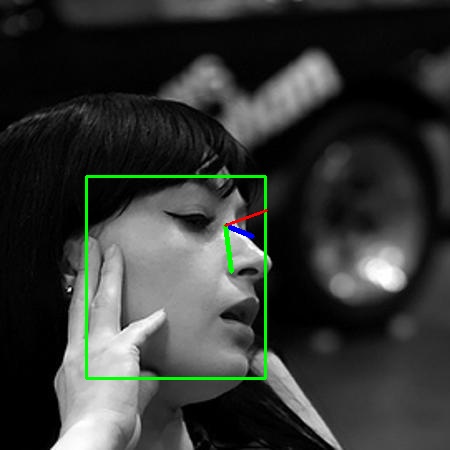}
\includegraphics[width=0.23\textwidth]{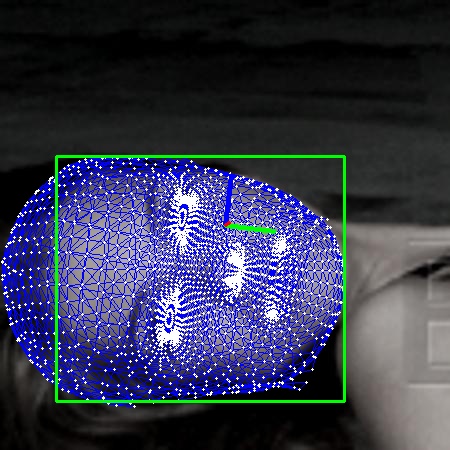}
\includegraphics[width=0.23\textwidth]{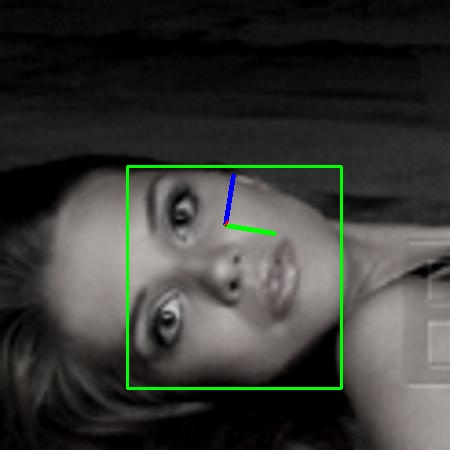}
\caption{\textbf{Qualitative Evaluation on AFLW.}}
\label{fig:aflw}
\end{figure*}

\begin{figure*}[htb]
\centering
\includegraphics[width=0.46\textwidth]{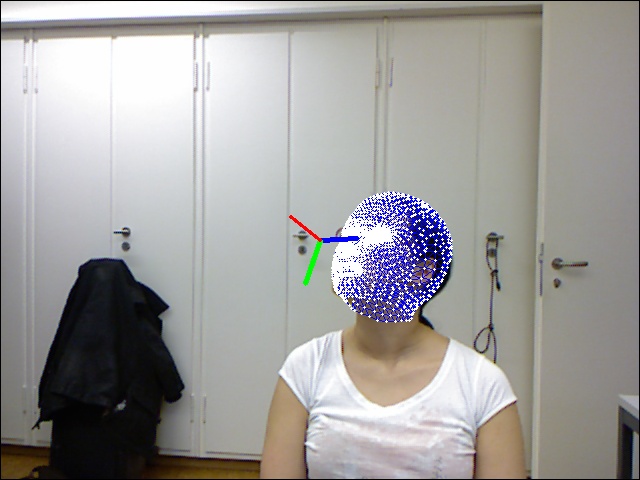}
\includegraphics[width=0.46\textwidth]{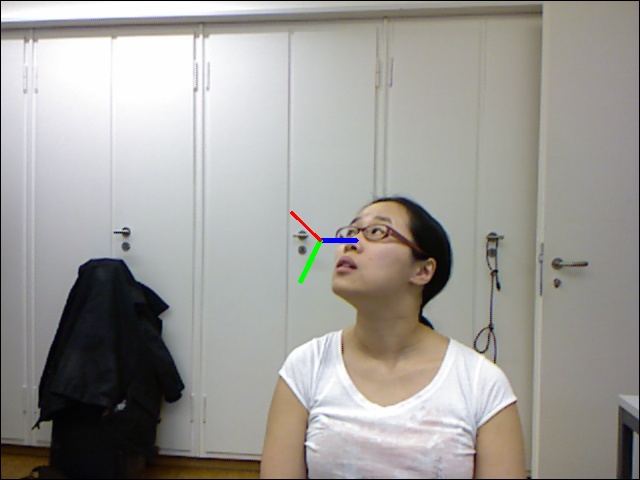}
\includegraphics[width=0.46\textwidth]{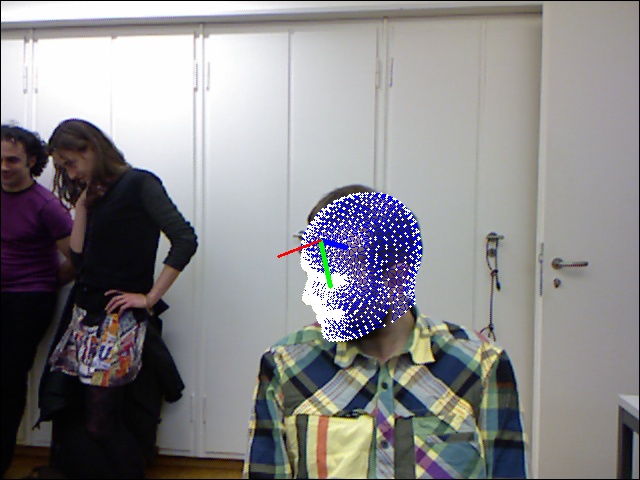}
\includegraphics[width=0.46\textwidth]{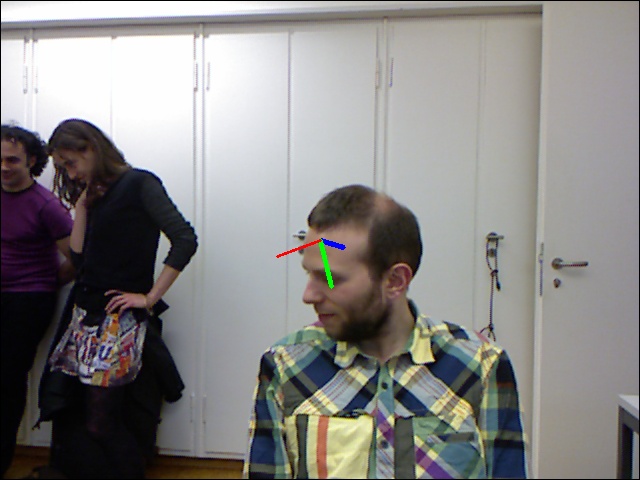}
\includegraphics[width=0.46\textwidth]{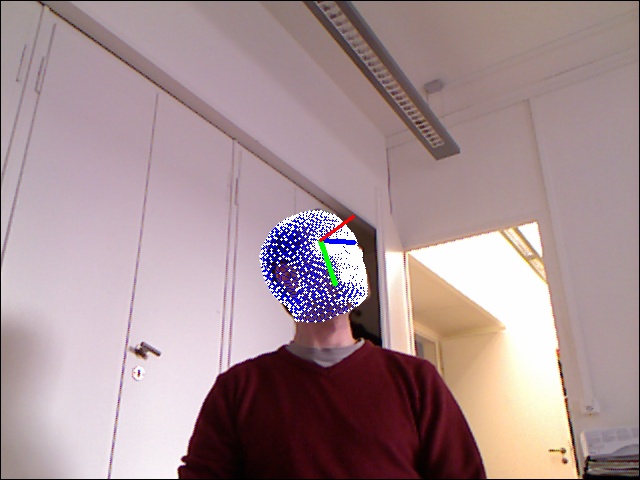}
\includegraphics[width=0.46\textwidth]{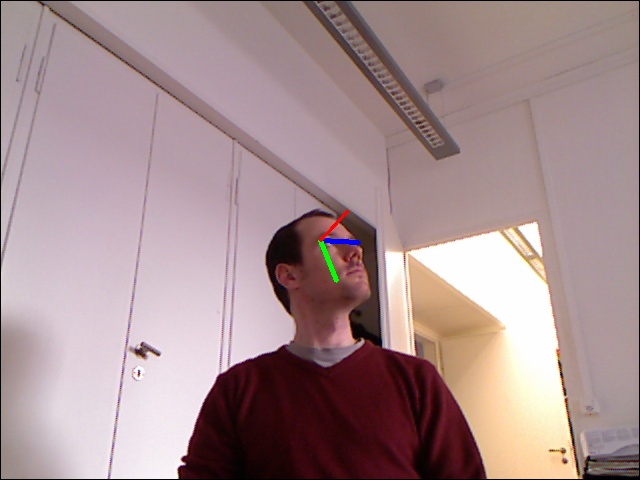}
\includegraphics[width=0.46\textwidth]{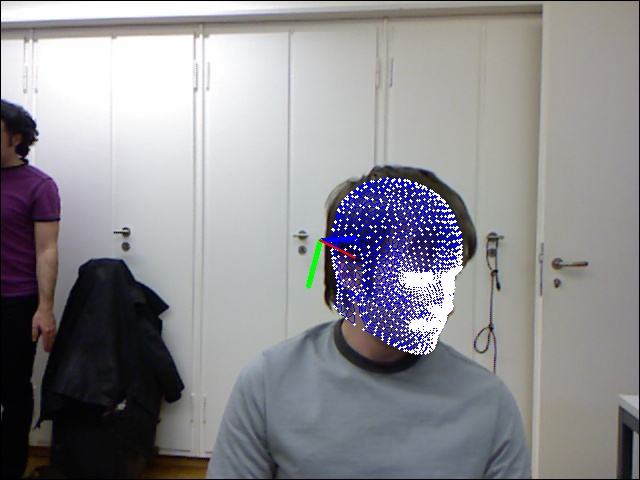}
\includegraphics[width=0.46\textwidth]{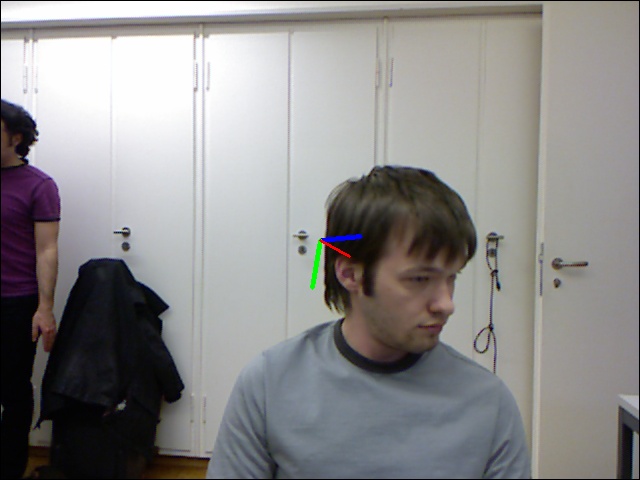}
\caption{\textbf{Qualitative Evaluation on BIWI.}}
\label{fig:biwi}
\end{figure*}

\begin{figure*}[htb]
\includegraphics[width=0.99\textwidth]{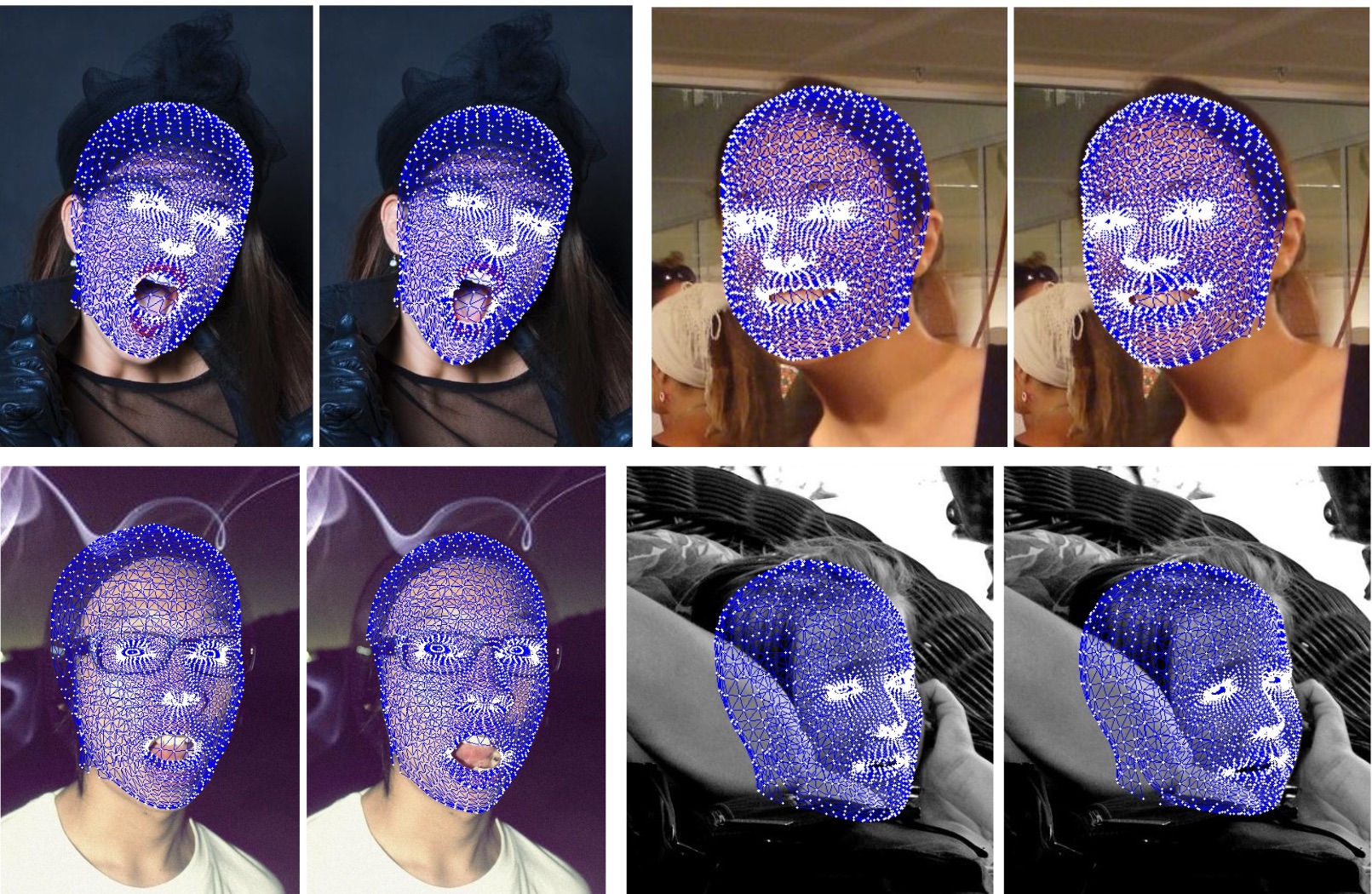}
\caption{\textbf{Qualitative Evaluation on DAD-3D}}
\label{fig:dad_qual}
\end{figure*}

We also include more visual results on AFLW \cite{aflw} \cref{fig:aflw}, BIWI \cite{biwi} \cref{fig:biwi}, DAD-3D \cite{dad3d} \cref{fig:dad_qual}, WIDER \cite{wider} \cref{fig:data_example} and FDDB \cite{fddb} benchmarks \cref{fig:fddb_qual}.

\section{Limitations and Broader Impact}

The dataset annotations are based on DAD-3D \cite{dad3d} so we don't aim to model neck, ears and eyeball vertices that are a part of the FLAME topology. The generation pipeline still can produce deformed small faces due to the limited resolution so we don't label and predict the 3D model parameters of the tiny faces. Also, the more advanced filtering methods and nsfw detection methods are a suitable venue for future explorations as on the large scale it is not feasible to guarantee the absolute correctness of the generated samples, even by adding the human evaluation into the process.
By leveraging synthetic data generated through diffusion models, we reduce the privacy, ethics, and safety issues in human subject research, as no real personal data is used so that privacy and ethical standards are upheld. Furthermore, the synthetic dataset's high resolution and detailed annotations provide a robust and versatile resource for developing and testing new models. This approach not only enhances the generalizability and accuracy of models trained on this data but also promotes ethical research practices by eliminating the need for real human subjects. The ability to generate large-scale synthetic datasets paves the way for safer and more inclusive research, free from the constraints and risks associated with real-world data collection. Thus, our work promotes ethical AI practices and sets a standard for future research in this area.

\section{Fail Cases}

\begin{figure*}[htb]
\includegraphics[width=0.95\textwidth]{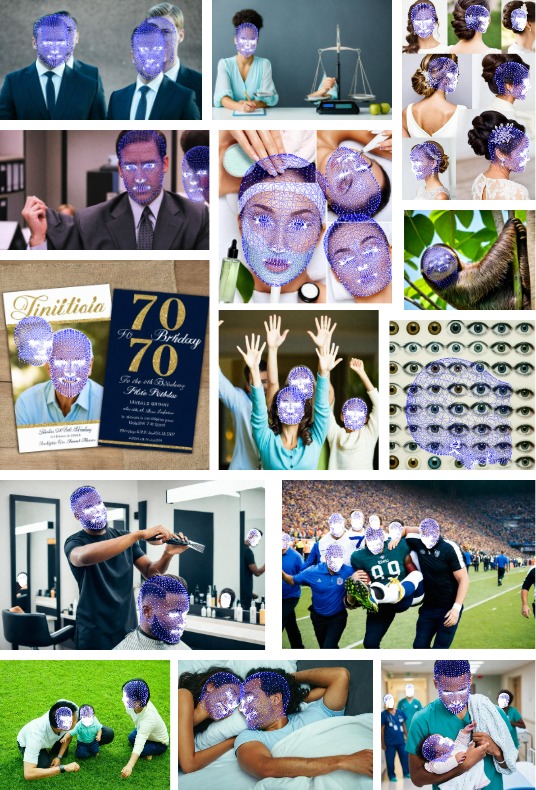}
\caption{\textbf{\method Dataset Fail Cases}}
\label{fig:fail_cases}
\end{figure*}

The dataset fail cases are presented in \cref{fig:fail_cases}. Typical failure cases are head detector failure (both FP and FN), misaligned mesh on out-of-distribution shapes and poses, severe occlusions, and deformed tiny faces in crowded scenes.

\end{document}